\pdfoutput=1
\documentclass{article}

\usepackage{PRIMEarxiv}
\usepackage{setspace}
\usepackage[colorlinks,linkcolor=black,anchorcolor=black,citecolor=black]{hyperref}
\usepackage{authblk}
\usepackage{longtable}
\usepackage{threeparttable}
\usepackage{graphicx}
\usepackage{subfigure}
\usepackage[font=small,labelfont=bf,tableposition=top]{caption}
\usepackage{booktabs}
\usepackage{xcolor}
\usepackage[noheads,nolists,notables]{endfloat}

\usepackage[utf8]{inputenc} 
\usepackage[T1]{fontenc}    
\usepackage{url}            
\usepackage{amsfonts}       
\usepackage{nicefrac}       
\usepackage{microtype}      
\usepackage{lipsum}
\usepackage{fancyhdr}       
\graphicspath{{media/}}     

\usepackage[font=normalsize]{caption}

\makeatletter
\def\UrlAlphabet{%
      \do\a\do\b\do\c\do\d\do\e\do\f\do\g\do\h\do\i\do\j%
      \do\k\do\l\do\m\do\n\do\o\do\p\do\q\do\r\do\s\do\t%
      \do\u\do\v\do\w\do\x\do\y\do\z\do\A\do\B\do\C\do\D%
      \do\E\do\F\do\G\do\H\do\I\do\J\do\K\do\L\do\M\do\N%
      \do\O\do\P\do\Q\do\R\do\S\do\T\do\U\do\V\do\W\do\X%
      \do\Y\do\Z}
\def\UrlDigits{\do\1\do\2\do\3\do\4\do\5\do\6\do\7\do\8\do\9\do\0}
\g@addto@macro{\UrlBreaks}{\UrlOrds}
\g@addto@macro{\UrlBreaks}{\UrlAlphabet}
\g@addto@macro{\UrlBreaks}{\UrlDigits}
\makeatother

\begin{document}


\title{Applications of knowledge graphs for the food science and industry
}
\date{}
%

\newcommand\blfootnote[1]{%
\begingroup
\renewcommand\thefootnote{}\footnote{#1}%
\addtocounter{footnote}{-1}%
\endgroup
}

\author[1,2]{Weiqing Min}
\author[1,2]{Chunlin Liu}
\author[3]{Leyi Xu}
\author[1,2,*]{Shuqiang Jiang}
\affil[1]{Key Lab of Intelligent Information Processing, Institute of Computing Technology, Chinese Academy of Sciences, Beijing, 100190, China}
\affil[2]{University of Chinese Academy of Sciences, Beijing, 100049, China}
\affil[3]{Soochow University, Suzhou, Jiangsu, 215021, China}

\blfootnote{* Correspondence: E-mail: sqjiang@ict.ac.cn.}


\maketitle



\begin{abstract}

The deployment of various networks (e.g., Internet of Things (IoT) and mobile networks), databases (e.g., nutrition tables and food compositional databases) and social media (e.g., Instagram and Twitter) generates huge amounts of food data, which present researchers with an unprecedented opportunity to study various problems and applications in the food science and industry via data-driven computational methods. However, these multi-source heterogeneous food data appear as information silos, leading to the difficulty in fully exploiting these food data. The knowledge graph provides a unified and standardized conceptual terminology in a structured form, and thus can effectively organize these food data to benefit various applications. In this review, we provide a brief introduction of knowledge graphs, and the evolution of food knowledge organization mainly from food ontology to food knowledge graphs. We then summarize seven representative applications of food knowledge graphs, such as new recipe development, diet-disease correlation discovery and personalized dietary recommendation. We also discuss future directions in this field, such as multimodal food knowledge graph construction and food knowledge graphs for human health.

\end{abstract}



\begin{spacing}{2.0}
\section{Introduction}

Food is critical to human life. It travels from the farm origin, through the growing, harvesting, packing, processing, transforming, production, transporting, distribution to consuming and disposing of food, forming the food system~\cite{FFS-Report2020}.
The production of huge volumes of multidisciplinary and heterogeneous food data~(e.g., nutrition composition table, health databases, food images, food ordering data and recipes) from the food system provides a basis for the development of Artificial Intelligence~(AI), making the digital technology an indispensable part of the food science and industry. Each stage from food processing to consuming in this system can be replaced with data-driven computational methods to prompt the development of food science and industry, such as the use of neural networks in modeling the food process~\cite{Raj2020Comprehensive,Qing2019Recent}, food quality assessment~\cite{Amani2020Current}, food object recognition and analysis~\cite{KNEZ2020Food,Marcus-VMVL-Patterns2020,LIU2021Efficient}, food authentication and traceability~\cite{Liang-AIS-CRFSN2020} and dietary assessment~\cite{thames2021nutrition5k,Larissa2021App}.

However, these food data are still not sufficiently utilized, and it is still hard to satisfy the demand for effective food data sharing, organization and traceability, which restricts the development of food science and technology. For example, in the food supply chains, the data from different food companies may be under different naming conventions, which restricts the aligning of food terms and the integration of different food data sources, making it harder to optimize the food supply system~\cite{Misra2020IoT}. In addition, more complex issues such as food contamination traceability~\cite{Qian2020Traceability} and exposure assessment involve data in multiple fields. They also require food systems to have abilities to integrate food data and organize food knowledge extracted from these multi-source heterogeneous data.

Therefore, there is one general agreement on the importance of organizing and integrating food data in the food science and industry. Only in this way, we can easily access and interchange food-relevant data all over the world, extract food-related information and organize food knowledge, which benefits different stakeholders, such as researchers, food manufacturers, food distributors,  retailers and consumers. For example, such a standardized knowledge organization system can facilitate the governance via more efficient knowledge access and utilization~\cite{Nicholas-Review-npj2018}, and food manufacturers and distributors can trace the processing and circulation of food commodities. All of them can make smarter decisions with the mentioned standardized knowledge organization system.

A key requirement for  standardization is to make  heterogeneous data from multiple sources interoperable. For that, the Internet of Food is proposed to help tackle this problem via defining one lingua franca~\cite{Nicholas-Review-npj2018}. Along with the changes in the form of data and the increasing volume of data, many types of lingua francas emerge with different ways of organizing food data. Considering the ontology describes more complex structures with arbitrary relations and restrictions between concepts~\cite{Igor-OKM-KIS2004}, different food ontologies have been developed, such as FoodOn ontology~\cite{Dooley-FoodOn-npj2019} and {ISO-FOOD} ontology~\cite{Eftimov-ISO-FOOD-FC2019}. Some communities, such as the Ontologies Community of Practice (CoP) have been created to  support the high-quality ontology development for agrifood research~\cite{Elizabeth-OCP-Patterns2020}. The food knowledge graph generally adopts the ontology as its schema to further model  more real-world instances and their relationships in a graph~\cite{paulheim2017knowledge,Haussmann-FoodKG-SIP2019}. It provides a unified and standardized conceptual terminology and their relations to link various information silos related to food, and can thus have a considerable impact in the food science and  industry. A range of applications include food safety (e.g., the traceback of  food contamination),  food allergy, chemical exposure and nutritional assessment, cooking and culinary use.

There have been some relevant reviews  on knowledge graphs from different perspectives~\cite{paulheim2017knowledge,Wang-KGE-TKDE2017,KRKG-Xiaojun-ESA2020,Shaoxiong-KG-arXiv2020,Hogan-KG-CSUR2021}. In contrast, this work seeks to provide a comprehensive review on  knowledge graphs in the food domain, namely food knowledge graphs, including the evolution from food ontology to knowledge graphs, their representative applications and prospects in the food science and industry.
\section{Knowledge Graph}
In this section,  the history of knowledge graphs is  briefly introduced and  how they are constructed, represented and used is also discussed. In order to better describe the development of knowledge graphs, commonly used terms are summarized in Table \ref{FTerms}.

\begin{table*}[!htbp]
    \centering
    \begin{threeparttable}[b]
	\footnotesize
    \renewcommand\arraystretch{1.5}
	\caption{A glossary of commonly used terms in knowledge graph}
    \label{FTerms}
	\begin{tabular}{p{60pt}p{370pt}}
\toprule
    Term &Description\\ \hline
    \textbf{Entity}& An entity can be a real-world object~(instance) or an abstract concept. Each entity is with a collection of attributes, and relations among it.\\
    \textbf{Relation}& Relation, also named entity description, refers to the interlinked description of entities. It should have formal semantics and support entities to form a graph.\\
    \textbf{RDF}& It is short for Resource Description Framework and is a uniform standard  to describe entities and relations in the form of Subject-Predicate-Object triples\tnote{a}.\\
    \textbf{RDFS}& RDFS refers to Resource Description Framework Schema, which extends RDF by adding common predefined vocabularies and supports constructing light-weighted ontology.\\
    \textbf{OWL}& It is short for Web Ontology Language, the W3C standard for defining ontologies. It provides the mechanisms for creating all the necessary components of an ontology: concepts, instances, and properties (or relations).\\
    \textbf{IRI}& IRI is short for the Internationalized Resource Identifier, an internet protocol standard used to identify and locate every entity and relation uniquely. Common identifiers like Uniform Resource Locator~(URL) and  Uniform Resource Identifier~(URI) are subsets of IRI.\\
    \textbf{Classification} &Classification is  one systematic arrangement in groups or categories according to established criteria\tnote{b}.\\
    \textbf{Taxonomy} & Taxonomy is a classification of things {in a hierarchical form}. It is usually a tree or a lattice that expresses subsumption relations (i.e., A subsumes B meaning that everything that is in A is also in B.) The fundamental difference between taxonomy and classification is that taxonomies describe relations between items while classification simply groups the items\tnote{c}.\\
    \textbf{Semantic Web technologies}&Semantic Web Technologies refer to all the technologies needed in the construction of the Semantic Web, including Hypertext Web technologies like IRI and XML, Standardized Semantic Web technologies for querying (SPARQL), description (RDF) and schema (RDFS/OWL), and those unrealized or unstandardized Semantic Web technologies~(like proof and trust layer for inferring and validation, and user interface for interaction). All of these technologies are combined to support a complete knowledge graph. These web technologies are hierarchical, and each type of web technology exploits the capabilities of the layers below.\\
    \textbf{Ontology} & An ontology is a description  of concepts and relations (e.g., synonymy and meronymy). The main difference between ontology  and taxonomy is that a taxonomy is an ontology in the form of a hierarchy. In many systems, ontologies and taxonomies work together.\\
    \textbf{Schema} & Schema usually means the technology that provides the standard, rules and principles for entities and their usage: they define all the classes and attributes that entity of each class should have. Ontology is usually used as the schema in the knowledge graph. \\
    \textbf{Semantic Network} & Semantic network consists of nodes and edges, where nodes represent entities and edges represent the relations. There is no standard for the values of nodes and edges.\\
    \textbf{Linked Data}& Linked Data is about using the Semantic Web Technologies to connect related data that are not previously linked and emphasizes the link creation between different datasets. Since datasets of the linked data are open access, it is also called linked open datasets.\\
\bottomrule
    \end{tabular}
    \begin{tablenotes}
        \item[a]  \url{https://www.w3.org/TR/PR-rdf-syntax/}
        \item[b]  \url{https://classroom.synonym.com/difference-between-classification-taxonomy-10074596.html}
        \item[c]  \url{https://www.obitko.com/tutorials/ontologies-semantic-web/specification-of-conceptualization.html}
    \end{tablenotes}
    \end{threeparttable}
\end{table*}

\subsection{Brief History of the Knowledge Graph}

\begin{figure}
\centering
\includegraphics[width=0.9\textwidth]{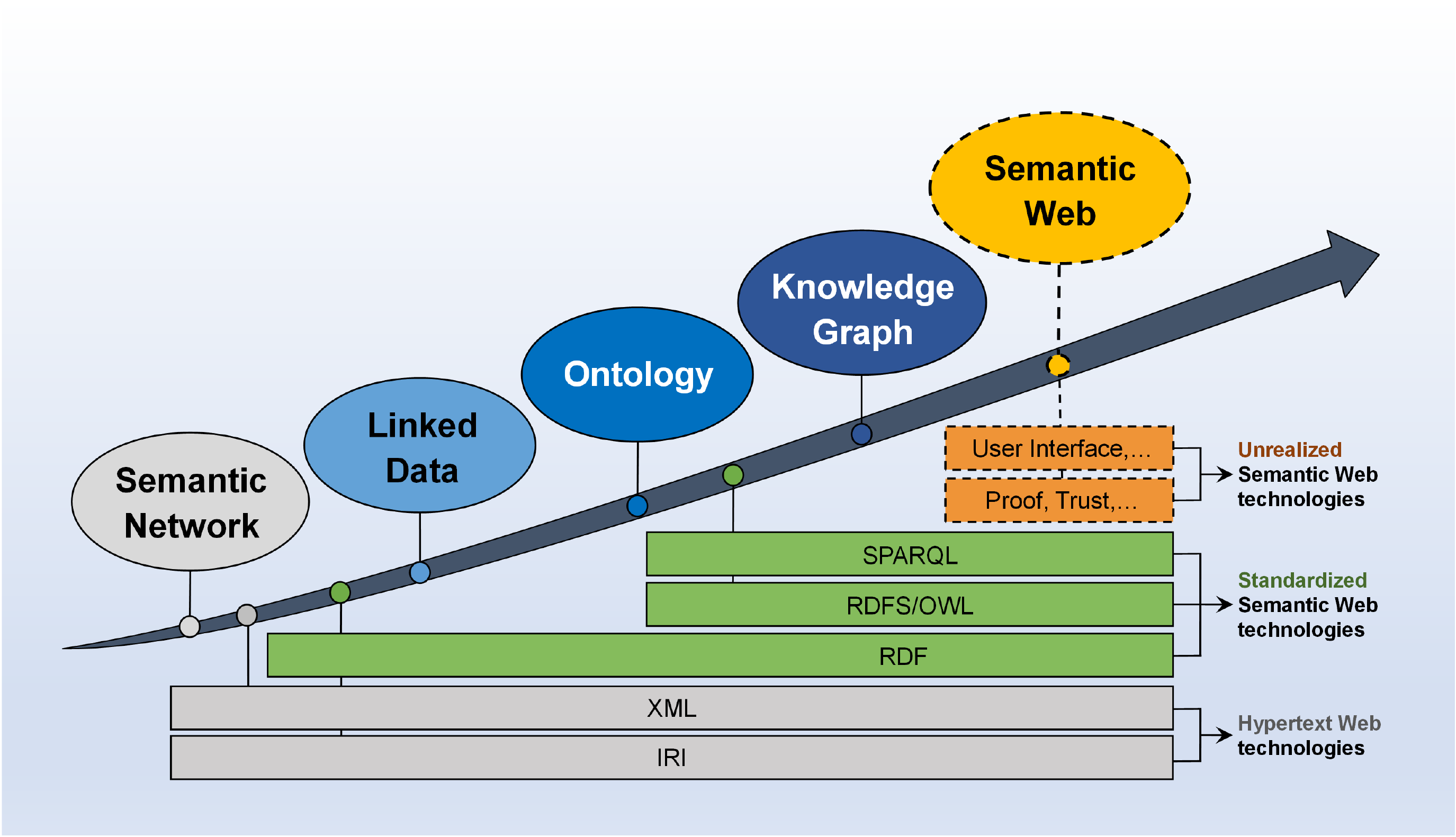}
\caption{The evolution of the knowledge graph.}
\label{FoodKG_hist_tech}
\end{figure}

The history of knowledge graphs and their related technologies are demonstrated in Figure ~\ref{FoodKG_hist_tech}. The graph is a type of sparse data structure that consists of nodes and edges, which is suitable to represent relations between objects. The idea of graph-based knowledge representation could trace back to the 1960s when the semantic network is first proposed as a form of knowledge representation~\cite{collins1969retrieval}. It uses the nodes to represent concepts and edges to represent relations between concepts in one graph. In semantic networks, there are no standards for the use of values of nodes and edges, which means the developers can freely define the nodes and their relations. Therefore, it is hard to integrate different semantic networks, making it difficult to apply semantic networks in practice.

Later, the Resource Description Framework~(RDF) is proposed to partially solve the problem of standards. RDF is developed by the World Wide Web Consortium (W3C) as a standard for describing web resources. The main data model of RDF is the subject-predicate-object triple expression, which indicates that the two entities (subject and object) are connected through a relation (predicate). These entities and relations generally use International Resource Identifiers (IRIs) as indexes in the RDF framework to address the difficulties in integrating data from different sources. This is because the same entity and relation have the same and unique IRI which has already been defined.

Based on RDF, Berners Lee proposes the concept of the Semantic Web, which is also known as Web 3.0. Semantic Web is a grand idea about the future Internet~\cite{berners1998semantic}. Its final goal is to make all the data on the Internet be published with semantics and linked with semantics to enable efficient and intelligent data querying, inference and understanding. In order to build the Semantic Web, W3C helps to build a technology stack called Semantic Web technologies, which could be involved in the construction of the Semantic Web~(e.g., RDF). Although Semantic Web remains largely unrealized, these technologies are widely used. Linked data is one of its implementations proposed in 2006~\cite{berners2006tabulator}, which publishes and interlinks datasets on the Internet using Semantic Web technologies.
Compared with the semantic network, linked data emphasizes links between web data and web resources. For example, elements of RDF triples of linked data are expected to be IRIs as much as possible, so that they can be unique and addressable on the Internet.

RDF still lacks the abstraction ability and can not describe or distinguish relations between entities, which affects knowledge understanding and inference. Thus, W3C successively proposes Resource Description Framework Schema~(RDFS) and Web Ontology Language~(OWL). RDFS and OWL extend RDF by adding common predefined vocabularies in the schema level so that they can represent abstract relations, like classes~(concepts), instances~(objects), subsets, and contains. The schema level is later separated to be the schema layer and is introduced to graph-based knowledge representation as a vocabulary and semantic specification. Many data models can be used as one schema layer, and ontology is the most widely used one. It's a knowledge specification, a formal explicit description of concepts within a certain domain, properties of each concept, and restrictions on facts. The aim of an ontology is to provide shared understanding to conceptual knowledge and give the definition to mutual relations between concepts~\cite{Gertjan-CSIO-AIM1995}, which makes semantic-based inference possible. Since RDFS and OWL provide good presentation capabilities and semantics supports, they are the main description language of the ontologies.

In 2012, Google proposes the term of knowledge graph~\cite{google2012kg}, which mainly describes real-world entities and their relations in a graphical representation, and  defines possible classes and relations of entities with the ontology as one schema~\cite{paulheim2017knowledge}.
It is synonymous with the knowledge base with a minor difference. A knowledge graph can be viewed as a graph when considering its graph structure. When we highlight formal semantics, it can be taken as a knowledge base for interpretation and inference over facts~\cite{Shaoxiong-KG-arXiv2020}. Currently, there is no  unifying definition of knowledge graphs. Herein we adopt the following definition: a knowledge graph is viewed as a multi-relational graph of data for conveying real-world knowledge, where nodes represent entities and edges represent different types of relations~\cite{Shaoxiong-KG-arXiv2020}. The focus of knowledge graphs is instances, while the ontology is  often used as the schema and plays a minor role in the knowledge graph. In general, the number of instance-level statements from knowledge graphs is far larger than that from the ontology~\cite{paulheim2017knowledge}.


\subsection{Knowledge Graph Construction, Representation, Reasoning and Applications}

\begin{figure}
\centering
\includegraphics[width=\textwidth]{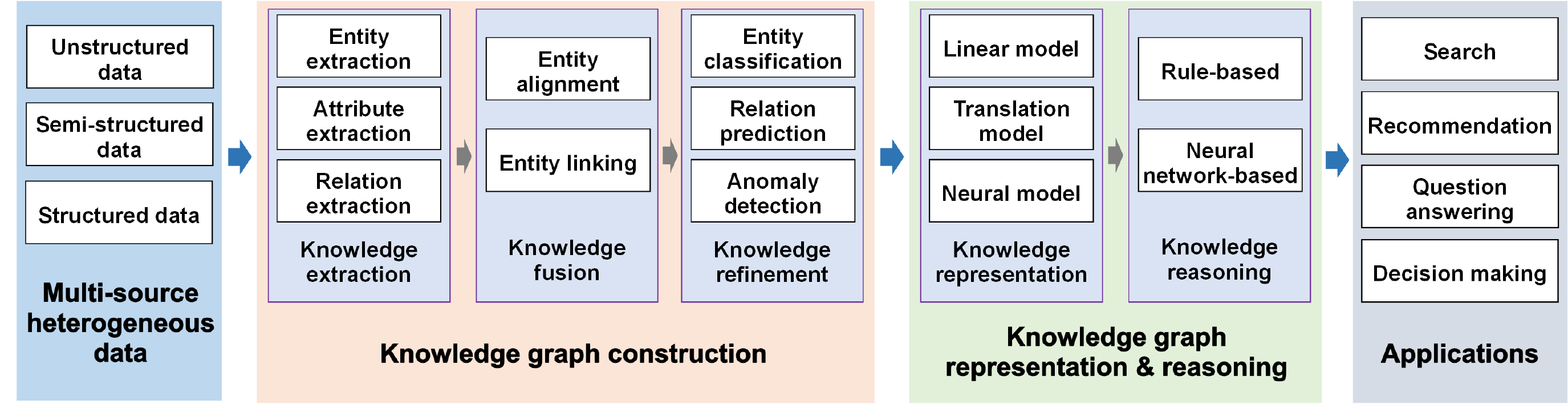}
\caption{Pipeline of knowledge graph construction, representation, reasoning and applications.}
\label{KG_framework}
\end{figure}

In order to explore knowledge graphs for applications, we first construct the knowledge graph.  Based on the constructed knowledge graph, effective representation  for knowledge graphs should be necessary to support further reasoning and applications, such as search and recommendation. The basic pipeline of knowledge graph construction, representation, reasoning and applications is summarized in Figure \ref{KG_framework}. More detailed and comprehensive introductions to knowledge graphs, such as knowledge graph creation tools and their more application examples can be referred to \cite{Hogan-KG-CSUR2021}.

\textbf{Construction}. Completeness, accuracy and data quality are three important factors that determine the usefulness of knowledge graphs and are influenced by the way knowledge graphs are constructed. Knowledge graphs can be constructed either manually or automatically~\cite{Nickel-RMLKG-IEEEPro2016}. Manual construction methods include curated ones (e.g., Cyc~\cite{Lenat-1995-CYC})  and collaborative ones (e.g., Wikidata~\cite{Denny2014wikidata}), where the former creates triples by a closed group of experts while the latter resorts to an open group of volunteers. Manually constructed knowledge graphs have little or no noisy facts. However, they require very large human efforts. As a result, auto-constructed  methods are explored and have become mainstream.

Auto-constructed methods can further be grouped into two types. The first one utilizes hand-crafted rules and learned rules to exploit semi-structured data, such as Wikipedia infoboxes, leading to larger, more highly accurate knowledge graphs such as DBpedia~\cite{Lehmann2014DBpedia}. This method can still guarantee high accuracy  of knowledge. However, semi-structured text still covers  a small fraction of the information stored on the Web, and these repositories are still far from complete. Hence the second approach is proposed to  extract facts from unstructured text using machine learning and natural language processing techniques. The knowledge vault~\cite{Luna-KV-KDD2014} is one representative  project in this category. In order to reduce the level of  "noise" in  extracted facts, a large body of research works are conducted, which mainly consists of  three components: knowledge extraction, knowledge fusion and knowledge refinement~\cite{paulheim2017knowledge}.

Knowledge extraction aims to acquire relevant entities, attributes and relations from various data sources. Information is collected and normalized, forming knowledge expression.  Considering there are multiple representations for one entity in many cases and inconsistency of triples extracted by multiple information extractors from multiple information sources,  knowledge fusion is one necessary step, and the main processes  include entity alignment and entity linking~\cite{Han-CEL-SIGIR2011}, where  entity alignment is the process to judge whether different entities refer to the same real-world object or not, and entity linking links the entities in text with the corresponding one in knowledge graphs. After initial  construction, different refinement methods, such as entity classification, relation prediction and anomaly detection~\cite{paulheim2017knowledge} are then utilized to improve the quality of constructed knowledge graph.

\textbf{Representation and Reasoning.} Effective representation learning for knowledge graph (namely, knowledge graph embedding) then should  be  explored based on the constructed knowledge graph. It can encode both entities and relations into a continuous low-dimensional vector space.  Different representation learning methods such as linear models, neural networks and translation methods are proposed~\cite{zhang2020transrhs,nayyeri20215,cenikj2021foodchem}. Based on learned feature representation, we can further conduct knowledge graph reasoning to identify errors and infer new conclusions from existing data. New relations among entities can also be derived through knowledge reasoning and in turn can be used to enrich the knowledge graphs.  Different reasoning methods  such as  rule-based reasoning and neural network-based reasoning are proposed~\cite{KRKG-Xiaojun-ESA2020}. Note that neural networks have been widely used  for  knowledge graph representation and reasoning for their powerful nonlinear fitting capability~\cite{Tay-MTNN-CIKM2017}.

\textbf{Applications.}  Knowledge graph representation and reasoning can  support various tasks, such as relation extraction and entity classification~\cite{Wang-KGE-TKDE2017} and real-world applications, such as Question-Answering(QA), information retrieval, and recommender systems. Here, we briefly discuss four critical use cases: search, recommendation, QA and  decision-making.

For \textbf{Search}, the knowledge graph can be used to understand  user's query intents to support semantic search, which  aims to not only find keywords, but to determine the intent and contextual meaning of the query words a person is using. Semantic search provides more meaningful search results by evaluating the search phrase and finding more relevant results. The knowledge graph enhances semantic search by providing more structured search results and better summaries. With the knowledge graph, the search engine  can summarize relevant content around that topic in the form of knowledge cards, including key facts for that particular thing. For example, when users search "apple cake", the content presented by knowledge cards includes various attribute information (e.g., cuisine, course, main ingredients) and other relevant information. In addition, it can expand the user's search results via the rich association of entities in the knowledge graph. For example, when the user searches for "apple cake", besides its basic information, semantic search can return its cooking recipes about them.

For \textbf{QA}, it has applications in a wide variety of fields such as chatbots. Answering questions using knowledge graphs adds a new dimension to these fields. As outlined by L. Hirschman and R. Gaizauskas, a knowledge graph based QA system involves answering a natural language question using the information stored in a knowledge graph~\cite{Hirschman-NLQA-NLE2001}. The input question is first translated into a formal query language and then this formal query is executed over the knowledge graph to fetch the answer. Such systems have been integrated into popular web search engines like Google and Bing Search as well as conversational assistants like Siri~\cite{Nilesh-NNA-CoRR2019}.

For \textbf{Recommendation}, the recommendation systems based on knowledge graph connect users and items, which can integrate multiple data sources to enrich semantic information. Implicit information can be obtained through reasoning techniques over knowledge graphs to improve recommendation accuracy. There are several typical cases for knowledge graph based recommendation, such as food recommendation~\cite{Haussmann-FoodKG-SIP2019}, movie recommendation and music recommendation~\cite{Oramas-SMR-TIST2017}. Knowledge graphs can benefit the recommendation from three aspects~\cite{Hongwei-RippleNet-CIKM2018}: (1) the knowledge graph can introduce the semantic relatedness among items to help find their latent connections and improve the precision of recommended items; (2) various types of relations in the knowledge graph is very helpful to extend a user's interests and increase the diversity of recommended results; (3) the knowledge graph can bring explainability to recommender systems via the connection between users' historical records and the recommended ones.

For  \textbf{Decision-making}, it is the act of choosing between possible solutions to a problem. Knowledge graphs stores expert knowledge from different domains to support high-complex decision-making. As one representative domain, knowledge graphs are actively used in the medical domain. When applied to medical knowledge graphs, reasoning on knowledge graphs can help doctors to diagnose disease and control errors to build a decision support system.

\section{Search Method}

We search for articles where food semantic data organization~(such as food linked data, food ontologies and food knowledge graphs) are proposed or utilized using the following electronic databases: 
IEEE Electronic Library (\url{ieeexplore.ieee.org}), ACM Digital Library (\url{dl.acm.org}),  Science Direct (\url{www.sciencedirect.com}), MEDLINE (PubMed, \url{pubmed.ncbi.nlm.nih.gov}) and Arxiv (\url{arxiv.org}). The following descriptors are used as a strategy for search in titles and abstracts: (food OR diet OR cook OR nutrition) AND ("linked data" OR ontology OR "knowledge graph" OR "semantic web" OR "semantic network"). 

Our search strategy is not restricted by publication year and language. We apply the following criteria for the inclusion of studies: (1) at least one food semantic data organization is proposed or utilized in research. (2) the food semantic data organization should be designed for food purposes specifically (such as cook, diet, recipe, health care and food production). (3) the construction or the usage of the food semantic data organization should be described in detail. The following exclusion criteria are applied: (1) the research is irrelevant to our topic (including not developed for food domains specifically). (2) no food semantic data organization is proposed or utilized in research.

\begin{figure}
\centering
\includegraphics[width=0.95\textwidth]{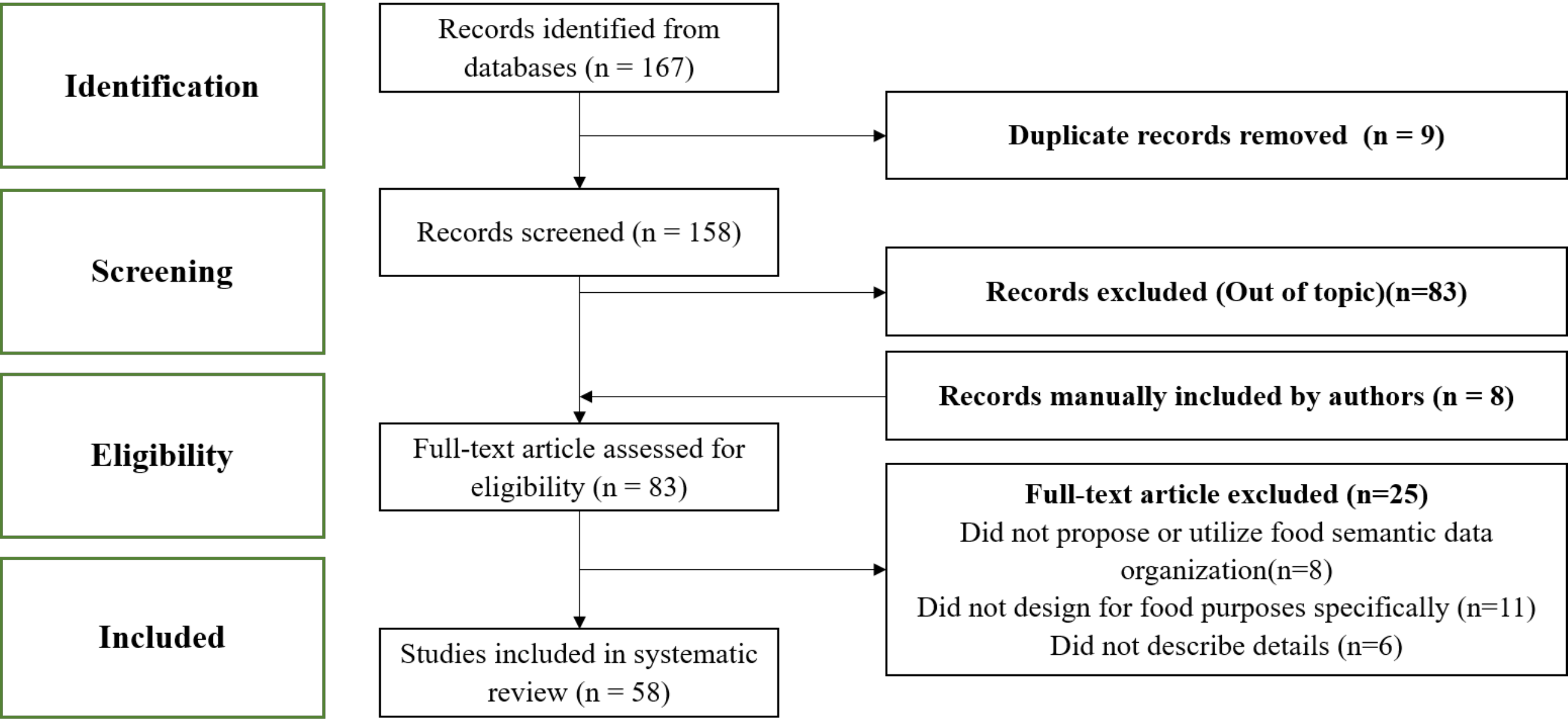}
\caption{Flowchart of the study selection process.}
\label{fchart}
\end{figure}

A total of 167 studies were identified through the searches in the databases. After the removal of 9 duplicate studies, 158 unique records remained, from which 83 studies were excluded based on their titles and abstracts because they were considered irrelevant. Later, the authors manually add several relevant studies that are not included in the above databases. Totally 83 studies are reviewed and evaluated in full for eligibility, and 58 meet all the criteria adopted for this review and are thus included, as the flowchart shown in Figure \ref{fchart}. All food ontologies and knowledge graphs selected in this article are summarized in Table \ref{FON} and \ref{FKG}, and these studies are organized chronologically by year of publication.

\section{Development of the Food Knowledge Graph}
Knowledge graphs allow for potentially interrelating arbitrary entities with each other from  various domains. When focusing on the field of food, they become food knowledge graphs.
Before delving into food knowledge graphs,  we first introduce the development of food ontology, since food ontology plays an important role in the development of food knowledge graphs. In addition, we also give some discussions on other forms of food knowledge organization, such as food-oriented linked data.

\subsection{Food Ontology}
Food ontology uses the shared terminology for types, properties and relations about food concepts, and thus can help tackle data harmonization problems that span  food-relevant domains. Table \ref{FON} summaries existing food ontologies from different aspects, where some  food ontologies have been introduced by previous work~\cite{Boulos2015Towards}.
Different food ontologies focus on different aspects of food and cover different sub-domains.
Particularly, we mainly divide them into the following four different types: (1) cooking and recipe ontologies, (2) nutrition and health ontologies, (3) other food sub-domain ontologies, and (4) more general and comprehensive food ontologies.

\begin{table*}[!htbp]
    \centering
    \begin{threeparttable}[b]
	\scriptsize
    \renewcommand\arraystretch{1.5}
	\caption{Summary on existing food ontologies}
    \label{FON}
	
	\begin{tabular}{p{150pt}p{18pt}p{108pt}p{145pt}}
\toprule
    Name&Year&Domain&Purpose\\
\midrule
    PIPS Food Ontology \cite{Cantais-FODC-SWeb2005}&2005&Food and nutrition&Providing food nutritional information\\
    Cooking Ontology \cite{Batista-OCcooking-2006}&2006&Food and cooking&Ontology construction research\\
    Food-Oriented Ontology-Driven System (FOODS) \cite{Snae-FOODS-ICDET2008}&2008&(Thailand) Food and nutrition & Food or menu planning for people with diabetes\\
    AGROVOC\tnote{a} \cite{Caracciolo-TMAPLD-MSR2011}& 2011& Agriculture, fisheries, forestry and food &Agricultural field terminology reference\\
    Edamam Food Ontology\tnote{b} &2012&Food, recipes, nutrition and healthy&Enabling food-related various applications, like healthy eating and cooking robots\\

    Food Track\&Trace Ontology (FTTO) \cite{TPizz-foodtrace-2013}& 2013 & Food supply chain & Supporting modeling of the food supply chain \\
    Open Food Facts\tnote{c} &2013& Packaged food product information& Food product comparison and search\\
    BBC Food Ontology\tnote{d}&2014&Food, recipes and diets&Recipe data publishment \\
    Taaable Cooking Ontology \cite{Cordier-Taaable-2014} & 2014 & Food, cooking and nutrition & Personalized cooking\\
    Unified Traveler and Nutrition ontology \cite{Karim-Travelers-2015} &2015&Food dishes and medicine&Healthy food recommendation\\
    FOod in Open Data Ontology \cite{Peroni-FOod-2016}&2015&General food&Creating linked open data datasets\\
    Food Ontology Knowledge Base~(from FoodWiki) \cite{Aelik-FoodWiki-SWJ2015}&2015&Packaged food& Building ontology-driven mobile safe food consumption system\\
    Food Product Ontology \cite{Maxim-FOODpedia-ESWC2015}&2016&Packaged food&(Russia) food products and domain data\\
    Ontology for Food Processing Experiment~(OFPE)\tnote{e} &2016&Food processing & Research on food processing \\
    Process and Observation Ontology($PO^2$) \cite{Ibanescu2016PO2}&2016&Food processing & Research on food production processes with data from different disciplines \\


    RICHIFIELDS Ontology \cite{Eftimov2018315}&2017&General food&Food-related integration, retrieval and updating\\
    Agri-Food Experiment Ontology (AFEO)\tnote{f} &2017& Viticultural practices and winemaking products & Research about food traceability and quality\\
    MEat Supply Chain Ontology~(MESCO) \cite{PIZZUTI2017MESCO} & 2017 & Food supply chain & Meat supply chain \\
    FoodOn Ontology \cite{Dooley-FoodOn-npj2019}&2018&Food sources, categories, products and other facets&Increasing global food traceability, quality control and data integration\\
    HeLiS \cite{Mauro-HeLiS-ISWC2018}&2018&Food and nutrition&Users' actions and behaviors monitoring\\
    Ontology for Nutritional Studies (ONS) \cite{Vitali-ONS-GeNutri2018} &2019&Food and nutrition&Nutritional studies\\
    ISO-FOOD \cite{Eftimov-ISO-FOOD-FC2019}&2019&Food and isotopic&Describing isotopic data within food science\\
    Food Safety Ontology \cite{Qin2019FSKG}&2019& Food safety&QA on food safety\\
    Food-Biomarker (FOBI) Ontology \cite{Caste2020FOBI}&2020&Food nutrition and metabolite&Food nutrition and metabolite research\\
    Supply Chain Traceability Ontology (SCT) \cite{Ameri2020Enabling}&2020&Food supply chain&Support agricultural food traceability\\
    Seafood Ontology \cite{sherimon2021modeling}&2021& Seafood & Seafood quality control\\
    Food Explanation Ontology (FEO) \cite{Padhiar2021SemanticMF}&2021& Food knowledge about recommendation and explanation &Providing users explanations for food recommendation\\
    Ontology of Fast Food Facts (OFFF) \cite{amith2021ontology} &2021& Food and nutrition &  Fast food nutritional data aggregation\\
\bottomrule
    \end{tabular}
	\begin{tablenotes}
        \item[a] The origin of AGROVOC can be traced back to the 1980s, and its linked data version is realized in 2011.
        \item[b] \url{https://www.edamam.com/}
        \item[c] \url{https://world.openfoodfacts.org/data}
        \item[d] \url{https://www.bbc.co.uk/ontologies/fo}
        \item[e] \url{http://agroportal.lirmm.fr/ontologies/OFPE}
        \item[f] \url{http://data.agroportal.lirmm.fr/ontologies/AFEO}
    \end{tablenotes}
    \end{threeparttable}
\end{table*}

\textbf{Cooking and recipe ontologies.} Taaable~\cite{Cordier-Taaable-2014}, Cooking ontology~\cite{Batista-OCcooking-2006}, Edamam food ontology and BBC food ontology are about cooking and recipes. Cooking ontology is one of the earliest cooking-oriented ontologies. It mainly contains actions, foods, recipes, and utensils. In the cooking ontology, one recipe is organized by phases of the cooking process, where each phase is a sequence of sorted tasks, and each task is composed by action and incorporates information about needed and produced ingredients and their duration time. Recipes also have their classification, ingredient lists, and required utensils. Cooking ontology aims to enrich cooking-oriented QA by being integrated into a dialogue system. Edamam food ontology aims to support creating a comprehensive and authoritative food knowledge base on cooking information. To do this, Edamam extracts the recipes from websites and maps these recipe terms to professional industry databases to eliminate duplicates and ambiguity. It has already supported search applications on mobile platforms and web pages for consumers to provide various food knowledge information such as ingredients, nutrition information and allergies~\cite{Aelik-FoodWiki-SWJ2015}.
In contrast, BBC food ontology is a simple lightweight ontology for publishing data about recipes, including the foods they are made from and the foods they create as well as the diets, menus, seasons, courses and occasions they may be suitable for. These ontologies facilitate cooking recipe based works, such as mining, retrieval and recommendation ~\cite{Min-YAWYE-TMM2018,Sajadmanesh-KC-arXiv2016}.

\textbf{Nutrition and health ontologies.} Some food ontologies focus on health and nutrition concepts, which allow them to help healthy advising and monitoring in various food applications. For example, Personalized Information Platform for Health and Life Services (PIPS) food ontology~\cite{Cantais-FODC-SWeb2005} provides nutritional advice for diabetic patients. It presents an abstract model of different types of foods with nutritional information, including the type, amount and recommended daily intake of nutrients, with a total of 177 classes, 53 properties and 632 instances. Similarly, Food-Oriented Ontology-Driven System (FOODS)~\cite{Snae-FOODS-ICDET2008} is also designed to provide diet advice for diabetic patients. In contrast, this ontology contains more aspects and concepts, like patients' personal situations and characteristics of foods (such as food specifics and flavors). Thus FOODS can provide more personalized and suitable diet recommendations for diabetic patients.

Different from the above food ontology serving special population{color{red}s}, the unified Traveler and Nutrition ontology~\cite{Karim-Travelers-2015} can support food recommendation to help general tourists make personalized food-related choices and develop a healthy food plan. This recommendation system is required to give recommendations by combining various factors from the food itself to the cultural requirements of tourists and regions of interest. Therefore, besides food nutrition, this food ontology also integrates various types of concepts from dishes, people and medical conditions to support more comprehensive dietary recommendation. HeLiS~\cite{Mauro-HeLiS-ISWC2018} is created to monitor both users' actions and their unhealthy behaviors by providing the representation of both food and physical activity domains. Besides covering concepts from activities to nutrients in foods, HeLiS also introduces the user concept, and it thus can associate the specific health-related events with people for health monitoring or further nutrient applications. In contrast, the FoodWiki ontology is developed for the packaged food products on market shelves. It collects the nutrition content and provides packaged food recommendation while avoiding the impact of unhealthy or allergic ingredients on consumers~\cite{Aelik-FoodWiki-SWJ2015}. Later, FoodWiki is further developed to build an ontology-driven mobile safe food consumption system for monitoring food intake~\cite{CelikErtugru-FoodWiki-SWJ2015}. In addition, there are some ontologies, such as Ontology for Nutritional Studies (ONS)~\cite{Vitali-ONS-GeNutri2018} and ontology of fast food facts (OFFF)~\cite{amith2021ontology} for food nutritional science study. For example, ONS is presented to facilitate the integration of different terminologies from different sub-disciplines in dietary and nutritional research and finally supports nutritional studies.

\textbf{Food safety ontologies.} Some ontologies are developed for the food safety domain, where food traceability is mainly considered. For example, Food Track\&Trace Ontology~(FTTO)~\cite{TPizz-foodtrace-2013,Teresa-FTTO-JFE2014}  is developed for food traceability. It contains representative food concepts involved in a supply chain and is able to integrate and connect the main features of the food traceability domain. The Supply Chain Traceability ontology (SCT) is for the agri-food supply chain where the form of critical tracking events is unified to support agriculture and food traceability from logistics to production lines~\cite{Ameri2020Enabling}.

Some ontologies are developed for specific food categories. Considering that food processing industries employ different quality control systems to check the quality of the seafood, developers create unique concepts and examples for seafood ontology~\cite{sherimon2021modeling}, such as various freezing in processing (blanched frozen, cooked frozen, and uncooked frozen). The MEat Supply Chain Ontology (MESCO) adapts the meat supply chain area~\cite{PIZZUTI2017MESCO}. In MESCO, the concepts in the meat supply chain are specially designed, so the attributes of different meat products are adjusted to adapt to their different processing procedures and safety traceability methods, and thus supporting better meat supply chain management. There are also some ontologies, such as the food safety ontology~\cite{Qin2019FSKG}, which focus on the public issue of food safety and is built to support a food safety knowledge graph. It crawls the unqualified food data from web resources to build the ontology. This ontology organizes concepts about food, food hazards, and food inspection items together, and then maps them to the Hazard Analysis and Critical Control Points~(HACCP) system for food production.

\textbf{Other food sub-domain ontologies.} There are also ontologies that focus on other food sub-fields to better promote the food science and industry. For example, Ontology for Food Processing Experiment~(OFPE)  can describe the transformation process from raw materials to products for food processing experiments. It includes different classes that represent products and operations during food transformation processes, which can be classified into four main concepts, i.e., Product, Operation, Attribute, and Observation. ISO-FOOD~\cite{Eftimov-ISO-FOOD-FC2019} is developed for sharing and researching isotope food data. In ISO-FOOD, the factors that are related to isotope are recorded as attributes and different sources of food data are unified and integrated under the standards from ISO-FOOD, so that the research and application of isotope in food chemistry can be promoted. Food-Biomarker (FOBI) ontology~\cite{Caste2020FOBI} is designed to integrate nutritional and metabolomic data to support nutritional research because nutrition research has a strong correlation with food intake evaluation and diet habits. Thus FOBI defines concepts and relations between both foods and metabolites. Its development improves the reusability of nutritional and nutrimetabolomic data.  Similar subdomain-oriented ontologies also include Process and Observation Ontology($PO^2$)\cite{Ibanescu2016PO2} and Agri-Food Experiment Ontology (AFEO) for food processing, Food Processing Chain Ontology (Onto-FP) for wine-making~\cite{muljarto2014ontology} and Food Explanation Ontology~(FEO) for generating the explanation for food recommendation~\cite{Padhiar2021SemanticMF}.

\begin{figure}
\centering
\includegraphics[width=0.9\textwidth]{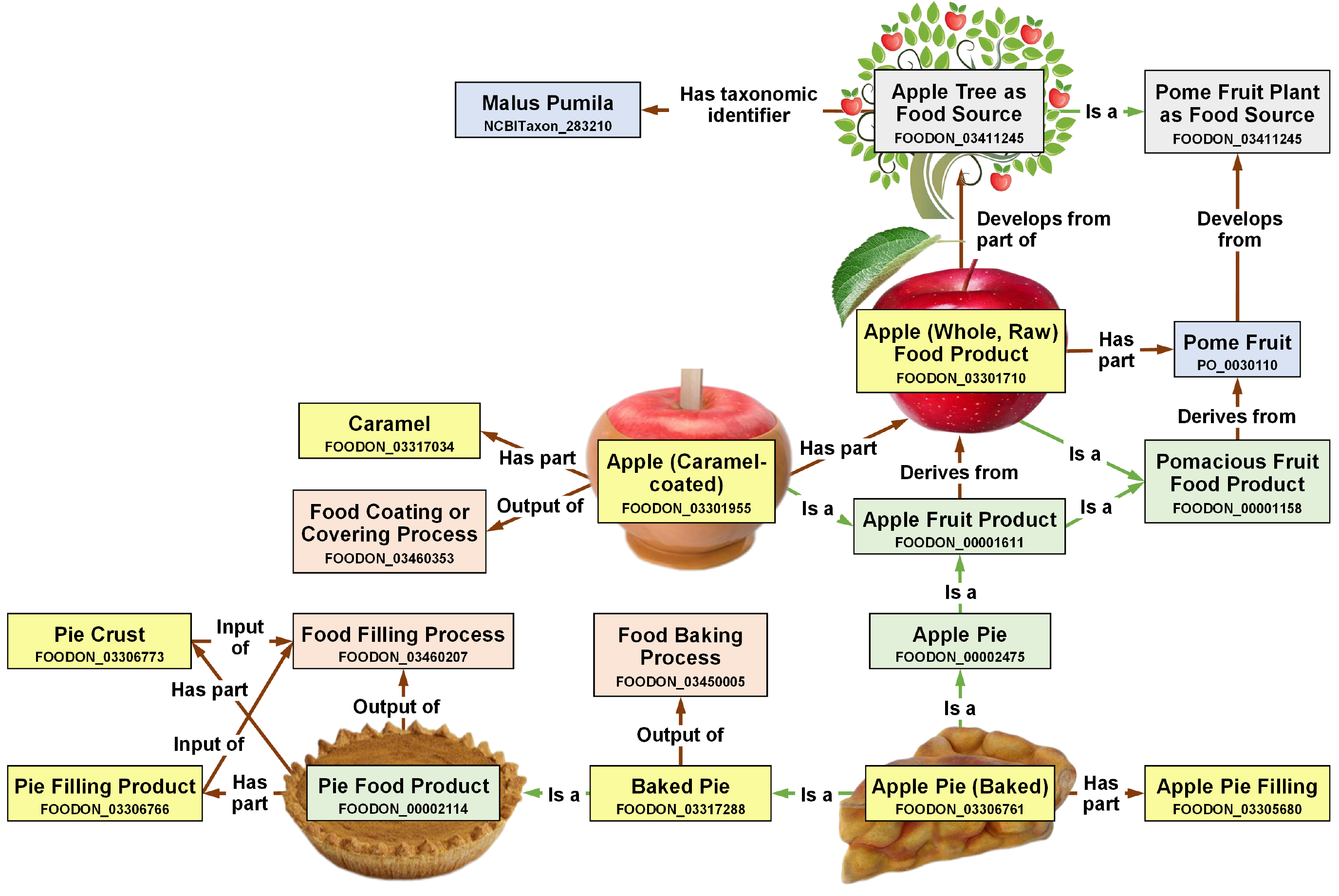}
\caption{A simplified structure of FoodOn~\cite{Dooley-FoodOn-npj2019}.}
\label{FoodOn_Img}
\end{figure}

\textbf{More general and comprehensive food ontologies.} There are also some ontologies with broader concepts, like FoodOn~\cite{Dooley-FoodOn-npj2019} and RICHIFIELDS ontology~\cite{Eftimov2018315}.
For example, FoodOn is an open-source, comprehensive food ontology resource composed of various term hierarchy facets from ingredients to packaging and cooking. FoodOn allows defining food product terms directly in the ontology and introduces relation descriptions like "has ingredient", "has part" and "derives from", which provides convenience for describing unique containment relations in food products. FoodOn acts as an interface with more food-specific domain ontologies, like food packaging, food nutrition and food processing. Its knowledge of both food and food processing is comprehensive enough to drive various applications, such as food safety, farm-to-fork traceability, risk management. Figure ~\ref{FoodOn_Img} shows a simplified example about apple food products in FoodOn. It describes the relations among food sources like apple tree and pome fruit plant, different kinds of food products like apple pie and caramel apple, and related food processes like food baking process and food coating or covering process. Moreover, considering that different food ontologies are developed for different application scenarios, these existing food ontologies can be integrated and reused to provide wider coverage of food concepts or serve for more general purposes. For example, FoodOntoMap is constructed to link these food ontologies~\cite{popovski2019foodontomap}, so that food concepts of different ontologies can be normalized by mapping them to a unified system. Thus, FoodOntoMap can be considered as a general food ontology  for further studies in different areas like diseases, human health or the environment.

In summary, food  ontologies formally describe food types, their properties and interrelations between food entities. However, these food ontologies generally lack detailed information about more food instances. For these reasons, food knowledge graphs are developed with  both food ontology and specific  food-relevant instances, where food ontology is generally considered as the schema.

\subsection{Food Knowledge Graph}

\begin{table*}[!htbp]
    \centering
    \begin{threeparttable}[b]
	\footnotesize
    \renewcommand\arraystretch{1.5}
	\caption{Summary on existing food knowledge graphs. "-" indicates unknown ontology, and "*" indicates that the food ontology is specialized constructed for the corresponding food knowledge graph.}
    \label{FKG}
	\begin{tabular}{p{180pt}llp{120pt}}
\toprule
    Name&Year&Ontology&Purpose\\
\midrule
    Knowledge Graph for the Food, Energy, and Water (FEW)\tnote{a}&2017&-&Data-driven research\\
    Chinese Food Knowledge Graph~\cite{ChiKnowledge}&2018&*&Healthy diet knowledge retrieval\\
    Foodbar Knowledge  Graph~\cite{Zulaika-EPCARFS-Proceedings2018}&2018&-&Small miniature bites or dishes cognitive gastroevaluation\\
    Healthy Diet Knowledge Graph~\cite{Huang2019Towards}&2019&*&Healthy diet management and recommendation\\
    Agricultural Knowledge Graph (AgriKG) \cite{Yuanzhe-AgriKG-DSAA2019} &2019&*&Agricultural entity retrieval and QA\\

    FoodKG~\cite{Steven-FoodKG-ISWC2019-1}&2019&WhatToMake ontology&Food recommendation\\
    Food Safety Knowledge Graph~\cite{Qin2019FSKG}&2019&Food safety ontology&QA system for the food safety domain \\
    Food Knowledge Graph with Dietary Factors and Associated Cardiovascular Disease~\cite{milanlouei2020systematic}&2020&-&Identifying dietary factors associated with Cardiovascular Disease \\
    Food Spot-check Knowledge Graph~\cite{Qin2020Question}&2020&Food safety ontology*&Food spot-check QA system\\
    Food Knowledge Graph (from World Food Atlas Project)~\cite{Rostami2021WFAP}&2021& FoodOn ontology & Supporting healthier and more enjoyable diets\\
    Recipe Knowledge Graph~(RcpKG)~\cite{LEI2021115708}&2021&-&Personalized recipe recommendation\\
\bottomrule
    \end{tabular}
    \begin{tablenotes}
        \item[a] \url{https://mospace.umsystem.edu/xmlui/handle/10355/62663}
    \end{tablenotes}
    \end{threeparttable}
\end{table*}

The proliferation of food-relevant instances, such as recipes and nutrition from various sources,  presents an opportunity for discovering and organizing food-related knowledge into the food knowledge graph.
We divide food knowledge graphs into four different types, mainly including (1) knowledge graphs about recipes, (2) knowledge graphs about nutrients and health, (3) knowledge graphs about food safety, and (4) general food knowledge graphs. Table \ref{FKG} lists constructed food knowledge graphs. 



\begin{figure}
\centering
\includegraphics[width=0.9\textwidth]{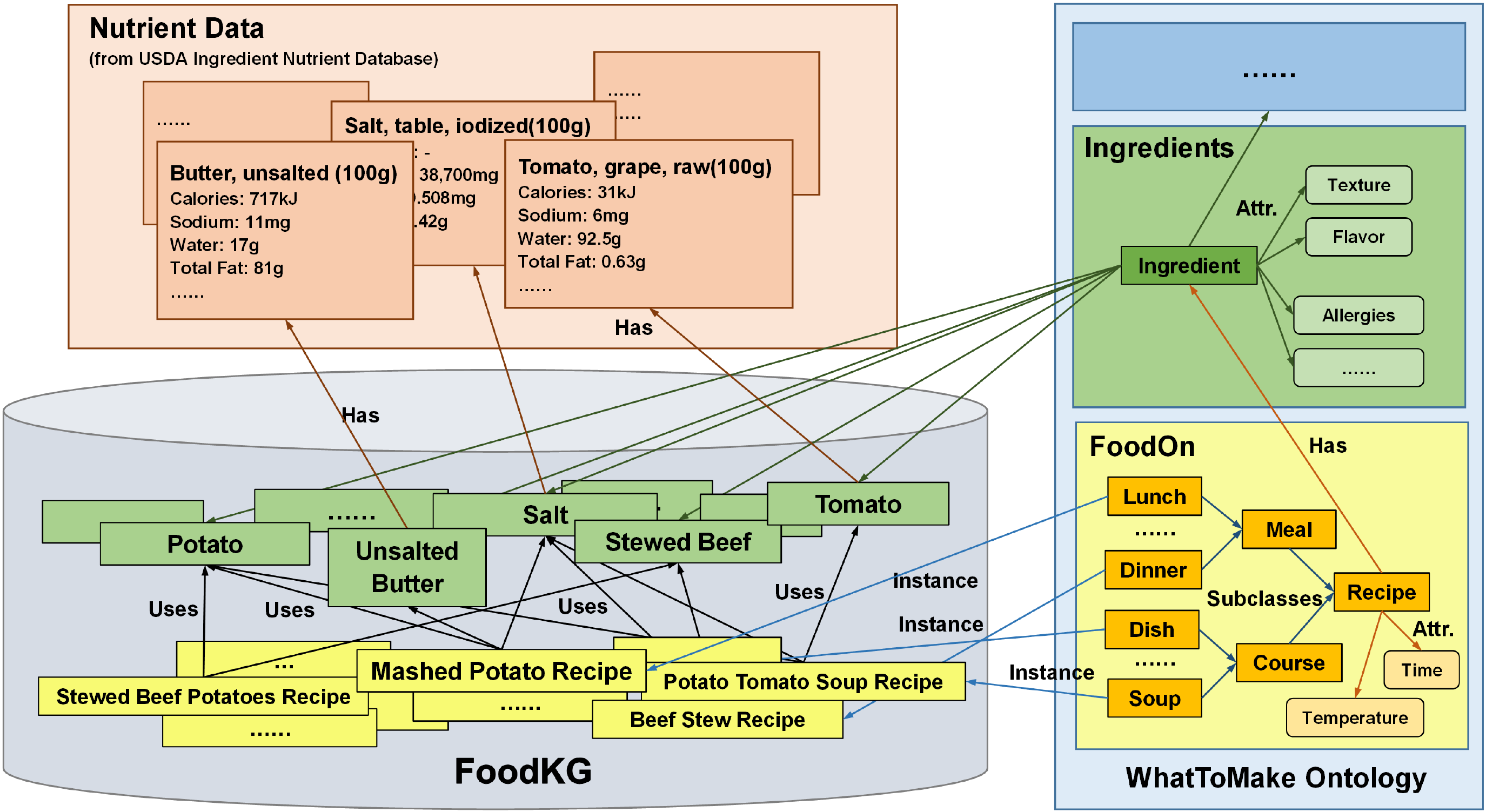}
\caption{The structure of FoodKG~\cite{Steven-FoodKG-ISWC2019-1}.}
\label{FoodKG_Img}
\end{figure}

\textbf{Knowledge graphs about recipes. } Some food knowledge graphs are mainly built based on recipe entities extracted from the crowdsourced consumer review sites, recipe-sharing websites and social media to support recipe-related applications. Foodbar knowledge graph~\cite{Zulaika-EPCARFS-Proceedings2018} contains more types of information, such as ratings and consumers' opinions from different restaurants and bars. It extracts above information from BEDCA and CookBook of Wikidata, and links to users, points of interest, cultural facts and so on. Based on this, Foodbar knowledge graph can be used for recommending miniature food according to the given user preference or providing food-relevant descriptive analytics services. Lei et al. further introduce social relationships into the food knowledge graph~\cite{LEI2021115708}. They construct a multi-modal and hierarchical recipe knowledge graph (RcpKG). In RcpKG, the users' demands are converted to nodes and modeled with specific hierarchical structures. Thus, it can link profiles of different users and give reliable recipe recommendations based on both personal preferences and social relationships. Its recipe data are from popular recipe websites~(e.g., Yummly and AllRecipes) and datasets~(e.g., Recipe1M+). 


\textbf{Knowledge graphs about nutrient and health.}
FoodKG is a large-scale and unified food knowledge graph, which brings together food ontologies, recipes, ingredients and nutritional data~\cite{Steven-FoodKG-ISWC2019-1}.  Particularly, as shown in Figure ~\ref{FoodKG_Img}, it integrates FoodOn into its WhatToMake ontology, and contains recipe and nutrient instances extracted from Recipe1M+~\cite{Salvador-LCME-TPAMI2021} and nutrient records from the United States Department of Agriculture~(USDA). Such food knowledge graph with more comprehensive recipe and nutrition information can support many applications, such as recipe recommendation, ingredient substitutions~\cite{shirai2020semantics} and QA~\cite{Steven-FoodKG-ISWC2019-1}.

Chinese food knowledge graph~\cite{ChiKnowledge} and healthy diet knowledge graph~\cite{Huang2019Towards} focus on food and medicine, especially ingredient and nutrient knowledge of Chinese food and Chinese medicine. Machine learning algorithms are used to extract information from Chinese health food websites and Chinese food composition table~\cite{yang2005chinese}, and their own ontology containing food-related concepts and relations are constructed, respectively. Chinese food knowledge graph basically supports semantic search, and the healthy diet knowledge graph~\cite{Huang2019Towards} further enable support more healthy diet applications, like QA and food recommendation.

Recently, there are some work on the relation between diet and disease. For example, Milanlouei et al. develop a knowledge graph of dietary factors associated with the cardiovascular disease~\cite{milanlouei2020systematic}. To create this knowledge graph, they collect and filter papers that study the association between dietary and cardiovascular complications from PubMed. They finally use 292 associations from 91 papers to construct the knowledge graph and environment-wide association study~(EWAS) approach is applied to discover relations between multi-types of diet and cardiovascular disease.

\textbf{Knowledge graphs about food safety.} Food safety knowledge graph~\cite{Qin2019FSKG} and food spot-check knowledge graph~\cite{Qin2020Question} mainly concern about food safety issues.
Food safety knowledge graph~\cite{Qin2019FSKG} contains the data of unqualified foods officially released in recent years from the Internet. Based on this knowledge graph, an intelligent food safety-oriented QA system is built to help people get the information of unqualified foods. Similarly, Qin et al. crawl food spot-check data from official websites of China national food quality supervision and inspection center and China food and drug administration, extract food spot-check information and construct a food spot-check knowledge graph~\cite{Qin2020Question}. A QA system is also provided based on the knowledge graph. 

\textbf{General food knowledge graphs.} Some food knowledge graphs cover more types of food-related knowledge from broader fields~\cite{Veron2020A}.
One of them is the knowledge graph for the food, energy, and water (FEW). It extracts the vast amount of available data from USDA, the national oceanic and atmospheric administration (NOAA), the United States geological survey (USGS), and the national drought mitigation center (NDMC) to support data-driven research for the food, energy, and water domain. Another one is agricultural knowledge graph~(AgriKG)~\cite{Yuanzhe-AgriKG-DSAA2019}, an agriculture domain-specific knowledge graph covering raw food materials and food products. Their agriculture data are extracted from some sources, like Wikidata, and the fragmented information is integrated for agriculture-relevant applications. In contrast, the World Food Atlas Project~\cite{Rostami2021WFAP} is built to include a wider range of food concepts. It is a project that can aggregate and unify food-related information from multiple offline and online sources in the world. To achieve this, researchers develop a food knowledge graph that uses FoodOn as its ontology to collect foods, ingredients and their relations from multiple sources. Later they develop the FoodLog Athl and the RecipeLog, two mobile applications for collecting diet records as dietary knowledge. Although still in the early stages, the combination of these two works will help explore the relations among food, culture and personal health, and promote regional food culture research.

Considering food knowledge graph construction needs a lot of laborious work, there are not many constructed food knowledge graphs in the academic field. In contrast, because of its vital importance in the food business~\cite{Natasha-InSKG-ACMCommun2019}, many companies such as Uber, Meituan and Yummly have constructed their food knowledge graphs to drive many products and make them more intelligent from different specific domains. For example, Uber Eats~\cite{uber2018kg} builds a food knowledge graph to enable food-related retrieval and recommendation. In this food knowledge graph, the nodes consist of different entities such as restaurants, cuisines and menu items, and  different relations are constructed as edges, such as the association between cuisines and location information. Edamam develops an extensive knowledge graph on food and cooking, including recipes, ingredients, nutrition information, measures and allergies. The goal of this food knowledge graph is to offer users multiple ways of searching to enable better food choices.

In order to effectively construct the food knowledge graph, one common method is to combine extractions from web content with domain knowledge from existing knowledge repositories. The semi-automatic way is usually adopted with both  machine learning methods and manual efforts. Generally, the first step is to construct the food ontology. One effective method  is to reuse existing food ontologies~\cite{HELMY20151071}. For example, FoodKG~\cite{Haussmann-FoodKG-SIP2019} adopts the ontology on food products from the FoodOn as its ontology. In some cases,  existing ontologies do not cover what is intended with the target project, and building one food ontology from scratch is thus necessary. The most widely used ontology construction method is to combine top-down and bottom-up approaches~\cite{Eftimov-ISO-FOOD-FC2019}, where the former starts with defining the classes for the more general concepts in the domain, and continues by defining the subclasses and the latter starts with a definition of more specific concepts in the domain as subclasses and continues by grouping these classes into more general concepts, such as Wine ontology~\cite{Virglio-OBP-2005}. As shown in Table \ref{FKG}, for food knowledge graphs like the Chinese food knowledge graph, AgriKG and food spot-check knowledge graph, their ontologies are specialized constructed from extracted data. After food ontology construction, more information on instance items, such as food entities and their relations~\cite{Popovski-FoodBase-Database2019,Popovski-FNE-Access2020,ChiKnowledge,DietHub-TFST-2021} should be extracted from various sources, and are added into the food ontology for food knowledge graph construction. There are also some food-oriented relation extraction models, like SAFFRON~\cite{cenikj2021saffron} for food-disease relation extraction and FoodChem~\cite{cenikj2021foodchem} for food-chemical relation extraction.

\subsection{Discussion}

Besides food ontology and knowledge graph, as mentioned in the history of the knowledge graph, linked data is also one way in organizing food knowledge. Some representative food linked data are also proposed, and they play their roles in the food science and industry. Among all of these work, AGROVOC~\cite{Caracciolo-TMAPLD-MSR2011}, FOODpedia~\cite{Maxim-FOODpedia-ESWC2015} and Open Food Facts are three representative linked data. For example, AGROVOC is considered as the largest food linked data about food and agriculture for the public, which has been coordinated by the Food and Agriculture Organization of the United Nations (FAO) since the early 1980s. AGROVOC introduces concepts to represent almost everything in food and agriculture and consists of over 39,500 concepts and 924,000 terms in up to 41 languages~(October 2021).
Besides, there are some thesauruses, which are not particularly designed for food domains, but involve terms of food and health classes, like food class in the DBpedia and "food and drinks" class in SNOMED Clinical Terms.


Sometimes, it is difficult to define what food linked data actually belongs to. Some food linked data datasets define the concepts and relations and use them to describe and represent the food domain. From this aspect, they can play the role of the food ontology. On the other hand, despite relatively limited application scenarios, food linked data may contain a large number of entities and organize them like the knowledge graph. For similar reasons, some food ontologies can be considered as food knowledge graphs, because they contain not only classes and their relations in a schema but also real-world instances, their properties and relations, according to the definition of knowledge graphs~\cite{paulheim2017knowledge}. For example, FoodWiki and FOODS contain not only the food ontology but also product instances, their properties and relations. This indicates that the boundary between food linked data, food knowledge graphs and food ontology is vague in some cases.

\section{Applications of Food Knowledge Graphs}
\begin{figure}
\centering
\includegraphics[width=0.80\textwidth]{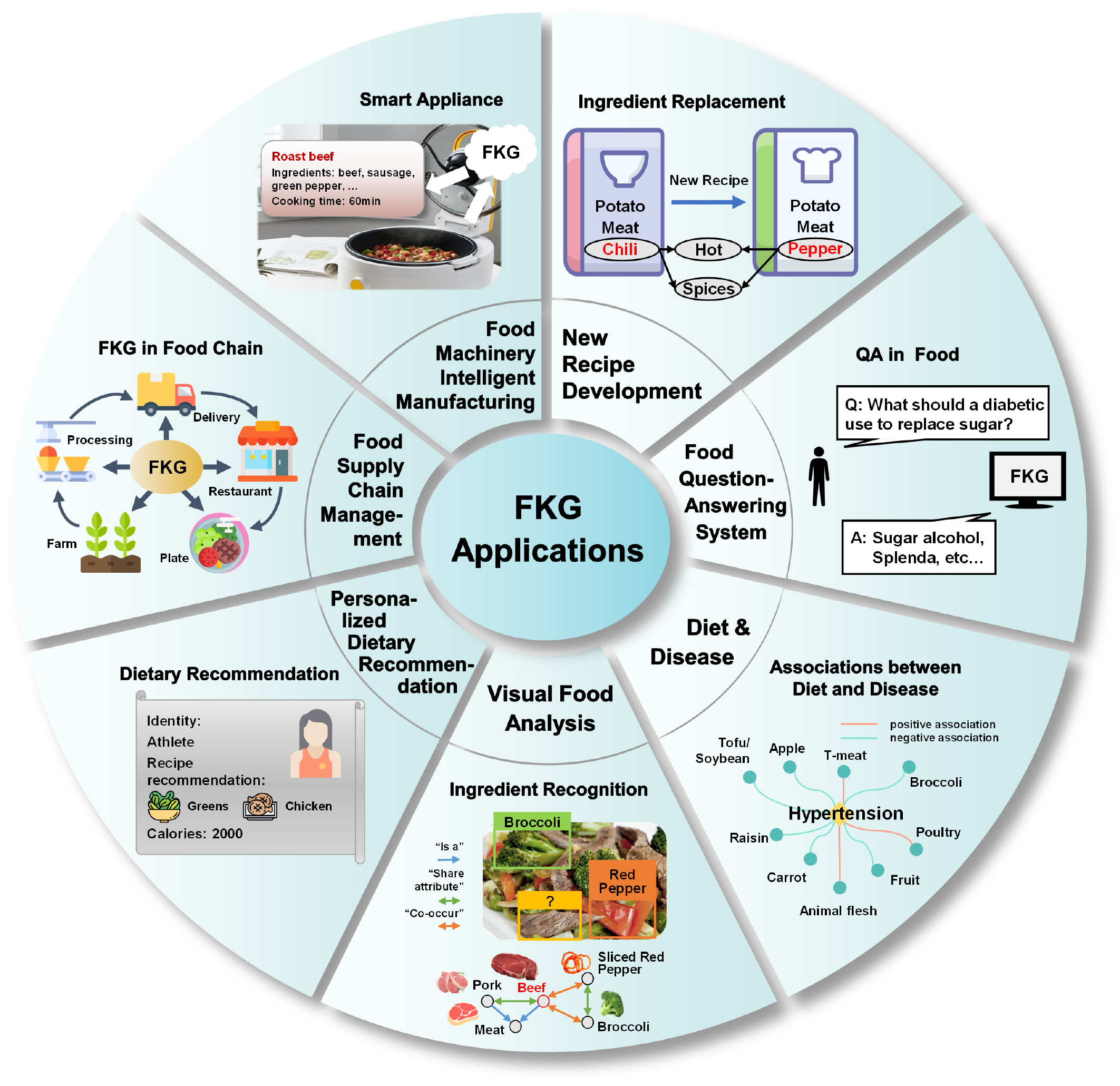}
\caption{Applications of food knowledge graphs~(FKG).}
\label{FoodKG_app}
\end{figure}

As illustrated in Figure ~\ref{FoodKG_app}, representative applications of food knowledge graphs in the food science and industry are identified and summarized  from the following seven aspects.  Considering food ontology is one important part of food knowledge graphs, we will discuss their applications in this section together.

\subsection{New Recipe Development}
The research and development of new food products is one important part of the food industry. Food knowledge graphs can be utilized to develop new products via effective knowledge organization and powerful inference ability on them . Developing new recipes is one representative application of food knowledge graphs. For existing recipes, we can resort to food knowledge graphs to find various alternative ingredients under requirements or develop new flavors~\cite{shirai2020semantics,Shirai-IISKGF-FAI2021}. Also we can develop novel culinary recipes, including not only their ingredient combinations, but also their ingredient proportions and time durations of each step via combining the constructed food knowledge graph and mathematical models~\cite{Pinel2015}.




First, we can use auto-infer abilities of food knowledge graphs to discover alternative ingredients. Shirai et al. develop a heuristic method to sort ingredient substitutions based on the similarity of the properties of ingredients and the similarity of latent semantic of ingredient names in FoodKG~\cite{Haussmann-FoodKG-SIP2019}. In their method, FoodKG is utilized as the latent semantics source in the form of word embedding since it includes abundant linked information about nutrition, ingredient and recipes, while Word2Vec~\cite{mikolov2013efficient} is utilized as the word embedding model. Considering that suitable substitute ingredients will have similar word embeddings, cosine similarity is used to measure and sort the best substitution ingredients. For the food industry, such a method can help discover ingredient substitutions for existing products, reduce food production costs. For consumers, such food knowledge graph-based ingredient substitution methods can give alternatives to specific recipes to meet their personalized needs. When integrating more comprehensive domain knowledge, the food knowledge graphs can give more personalized alternative ingredients based on more factors, including not only ingredients but also health indexes like glycemic index and glycemic load.

The food knowledge graph can also facilitate the development of new recipes. Generally, special flavors of foods come from the mixing or the interaction of several food components during cooking, while the specific mechanism inside may not be explored clearly. However, correlations of food components can be discovered statistically with co-occurrence information. Considering that the food knowledge graph can organize recipes and chemical components in a similar way, food manufacturers can use food knowledge graphs to discover well-matched latent ingredient combinations. Ahn et al. introduce a flavor network to capture the shared flavor compounds in various ingredients~\cite{Ahn-FNFP-SciRe2011}. They construct a bipartite network to link about 400 ingredients and over 1,000 flavor compounds, and then project it to a flavor network, where ingredients sharing the same flavor compounds are connected, and the weight of each link depends on the number of their shared flavor compounds. Then they use recipes from American repositories to analyze ingredient combinations in different regions, including popular and unpopular ones. Some ingredient combinations may only be popular in some regions, or combinations are feasible but not being tried yet. Applying these ingredient combinations to existing recipes is an effective way to develop new recipes, which may bring new flavors and genres.

In addition, the knowledge graph can also be used directly to develop new recipes. For example, Pinel et al. construct a food knowledge graph to organize data of recipes and ingredients, and generate recipes that fit the requirements of users based on the constructed knowledge graph~\cite{Pinel2015}. Later their algorithm selects the best ingredient combination and proportions by novelty and quality evaluators. Recipe steps are then generated using a subgraph composition algorithm, and the time duration of each step is estimated from known complete recipes.


Besides the above-mentioned applications, food knowledge graphs also show considerable prospects in more aspects of improving recipes. For example, considering that most foods already exist when the knowledge graphs are developed, it cannot be ignored that there are unhealthy (even hazardous) ingredients or pairing existing in these foods. Linked with toxicology knowledge, a food knowledge graph can assess the toxicity risk of the specific recipe from its components~\cite{Davis2016Comparative,Wishart2006Drugbank}. When food processing knowledge is integrated, it can even assess the potential risk of the specific recipe during the processing by its reasoning and advise substitute ingredients or substitute processing steps.

All of these above-mentioned works show that food knowledge graphs can make new recipe development possible, and exploring food knowledge graphs to develop new recipes provides a way of new recipe development with higher efficiency and reliability in the food industry.

\subsection{Food Question-Answering System}
Question-Answering (QA) system via food knowledge graphs can help people analyze the information and potential problems, and answer food-relevant questions about different food sub-domains, such as nutrition and disease, and food safety. For example,  a diabetic often asks questions like, "How can I increase the fiber content of this cake?" A person with lactose intolerance may ask "What can I substitute for milk in chocolate cake?" Answering these questions is not possible from general knowledge graphs for the incompleteness of domain knowledge.  Food knowledge graphs can be developed to support  natural language QA based on different categories of questions about recipes and nutrition, such as simple queries for nutritional information, comparisons of nutrients from different foods, and constraint-based queries to find recipes matching certain criteria~\cite{Steven-FoodKG-ISWC2019-1}.  Food knowledge graph-based QA systems can also describe recipes, nutrients in foods and the interaction between nutrients and prescribed drugs, disease and general health  to  satisfy users' specific information needs. For example, cooking QA~\cite{Riyanka-CookingQA-SIP2017,Semih-RecipeQA-EMNLP2018} is intended to  satisfy the user's information need in the cooking domain, and  is helpful to  people, such as housewives and nutritionists. FoodKG~\cite{Haussmann-FoodKG-SIP2019} organizes diets, nutrients and food types together, which can be leveraged by a QA system in the food field. It takes natural language questions as the input, and generates answers from the information stored in FoodKG. The questions it can answer can be roughly divided into three categories: simple questions, which directly ask about the ingredients of a certain food; comparative questions, where given some conditions, the system selects more suitable food; restricted questions, where given restrictions on ingredients or types of food, the system provides qualified food. When there comes a question, the system decides the question style, detects the mentioned topic entity, and then use a KBQA model to retrieve answers from FoodKG. The system is also enriched by user preferences to improve personalized question answering.

In addition, some question-answering systems via food knowledge graphs have been developed for the issue of food safety. For example, Qin et al. construct a food QA system based on the food safety knowledge graph which collects officially published food data from the Internet~\cite{Qin2019FSKG}. In their QA system, users’ questions are first parsed, and every food and attribute will be mapped to entities and relations in the food safety knowledge graph. Later questions in natural language will be converted to SPARQL query statements by template matching so that questions related to food safety can be understood and answered by the machine. When food data and templates change, such workflow can also be applied to answer different types of questions. For example, food spot-check knowledge graph~\cite{Qin2020Question} is utilized in a similar way to provide food spot-check data QA.

\subsection{Diet-Disease Correlation Discovery}
The research on diet, disease and their correlation modeling is always one important aspect in food science and nutrition. It has already been proved that there are inevitable connections between chronic diseases and certain diet styles~\cite{woodside2013fruit,Ashkan2019Health,Zhao2020Dietary}. Some studies also show that diseases without effective treatments (like neurodegenerative diseases) are associated with certain foods, which provides potential opportunities to prevent diseases or delay disease progression~\cite{Joseph2009Nutrition}. This is because we can build connections among diseases, diets, food, raw food materials and chemical components via constructing food knowledge graphs and then conduct deeper analysis for their correlations~\cite{milanlouei2020systematic}.

Jensen et al. build a system called NutriChem, a resource covering the broad molecular content of food, collecting exhaustive resources on the health benefits associated with specific dietary interventions~\cite{Jensen2014NutriChem}. NutriChem contains 18,478 pairs of 1,772 plant-based foods and 7,898 phytochemicals, and 6,242 pairs of 1,066 plant-based foods and 751 diseases. In addition, it includes predicted associations for 548 phytochemicals and 252 diseases. All of these data are generated by mining 21 million MEDLINE abstracts for information that links plant-based foods with their small molecule components and human disease phenotypes. To organize these data, they introduce an ontology that integrates the taxonomy from NCBI taxonomy,  the Plant For A Future~(PFAF) and the Danish Food Composition Databank. The relations in the ontology are built using Fisher’s exact test. NutriChem allows us to integrate established relations among food, compounds and diseases in a more comprehensive way. Therefore, we can easily understand the role of certain types of foods, and even infer which types of food is harmful or beneficial to certain food. This provides a foundation for understanding mechanistically the consequences of eating behaviors on health.

Nian et al. investigate relations between food and neurodegenerative diseases in a similar way~\cite{nian2021knowledge}. They collect biomedical annotations from over 4,000 publications and create the knowledge graph. Later the node2vec algorithm is used to train graph embeddings for clustering similar concepts and distinguishing different concepts. In the constructed knowledge graph, diseases nodes and diets nodes will be connected if they are relative, and their weights are determined according to the strength of the relevance. Based on this, they found that some food-related species and chemicals coming from the diet have a strong impact on neurodegenerative diseases. Similar work includes the biochemical knowledge graph,  a comprehensive source of knowledge for integrating biochemical knowledge and accelerating discovery in biochemical sciences, whose information is extracted and mined from biochemical literature~\cite{manica2019information}. Some online platforms, such as DietRx~(\url{cosylab.iiitd.edu.in/dietrx/}) can also collect the food-disease associations from MEDLINE abstracts, which can be used for exploring the interrelationships among food, chemicals, diseases and genetic mechanisms.

These leading-edge studies have proved the feasibility of the food knowledge graphs discovering the food-disease interaction. Thus, constructing a knowledge graph with diseases and food composition can be expected to analyze more general chemical component-disease relationships, generate novel insights, and even explore potential disease prevention strategies by further designing certain diets.


\subsection{Visual Food Analysis}
Rapidly and reliably detecting and analyzing food product quality and safety  (e.g., meat products, cereal products, fruits and vegetables) in one non-destructive way is significant for the food industry. Along with the development of AI, AI-augmented food analysis has become a new trend of food analysis~\cite{zhou2019application}. They use machine learning algorithms to process data from sensors (like spectral and chromatographic data). Vision-based food analysis  from the image sensor is usually considered for its non-destructive nature. Among all visual food analysis methods, visual food recognition is one basic task. Automatic food recognition can replace the manual grading process and quality detection~\cite{Qing2019Recent}, and  can also work as one basic step for various applications, like food log system and suggester system~\cite{Kiyoharu-FoodLog-CHSC2020,KNEZ2020Food}.

Knowledge graphs can be vectorized by machine learning algorithms to support visual food object recognition. Rich knowledge from food knowledge graphs, such as ingredients and their relations, have been explored to improve the performance of visual food recognition~\cite{Chen-ZSIR-AAAI2020,Min-IGCMAN-ACMMM2019,Min-MSMVFA-TIP2019}. For example, Chen et al. leverage multiple relations among ingredients for ingredient recognition~\cite{Chen-ZSIR-AAAI2020}. They construct a multi-relational knowledge graph to describe the ingredient relations and develop a graph model called multi-relational graph convolutional network~(mRGCN) for zero-shot ingredient recognition from the dish, namely, recognizing ingredients that the model hasn't seen. In mRGCN, the food knowledge graph is introduced as prior knowledge because it contains a large amount of recipe data, which provides the probability of coexistence between ingredients and the probability of food containing a certain ingredient. By ingredient recognition enhanced by food knowledge graphs, mRGCN can finally predict the dish category. There is a performance improvement of 9.7\% when introducing the ingredient knowledge graph and mRGCN achieves a 24.2\% top-1 hit for unseen ingredients in the VIREO Food-172 dataset, where top-1 hit measures the percentage of the most possible predictions that match the ground-truth labels. Based on the recognized dish type, we can also further resort to the food knowledge graph to obtain more detailed information about the recognized dish type~(such as properties, macronutrients and ingredients) to realize automatic dietary assessment~\cite{mezgec2017nutrinet,mezgec2019mixed}.

So far there is no work that uses food knowledge graphs to handle more complex food visual analysis tasks like food object detection and segmentation, where compared with food recognition, food detection additionally provided the localization for the recognized food item, and food segmentation is one process of assigning food labels to every pixel in one food image. However, it is proved that the object detection framework can integrate external knowledge from a knowledge graph to improve its performance because some combinations of objects are more common than others~\cite{FangKLTC17}. Image segmentation  augmented by knowledge has also been implemented in medical scenarios~\cite{DR2009Fuzzy,CHEN2012591}. We believe similar methods can also be applied in food scenes to improve the performance of complex food visual analysis tasks. For example, we can use the food knowledge graph to enhance the performance of food segmentation, which can further help the dietary assessment~\cite{s20154283,9091215}.
\subsection{Personalized Dietary Recommendation}
Personalized food and nutrition is gaining more and more attention in food science and other relevant domains~\cite{ueland2020perspectives,kirk2021precision}. They aim to use comprehensive personal information about individuals (e.g., dietary pattern and gut microbiota) for personalized dietary advice or recommendation, which is more suitable than generic advice. However, food recommendation can be a daunting task partly because of the problem of information silos across multiple sources with large amounts of food and nutrition data. In addition, different from other types of recommendation, food recommendation should take many nutritional parameters into consideration, such as caloric, different macronutrient and micronutrient intake.


A natural solution to this problem is to provide an intelligent food recommender system based on the food knowledge graph. Food knowledge graphs provide formal, uniform and shareable representations about food. They can benefit from different aspects, such as improving the precision of recommended items, increasing the diversity of recommended items and bringing better explainability~\cite{Hongwei-RippleNet-CIKM2018}. When it is organized with the personal health knowledge graph, it can further give a personalized dietary recommendation based on food knowledge graphs. This can benefit different people, such as diabetics~\cite{Chang-ontological-2008}, weightlifting athletes~\cite{Tumnark2013OntologyBasedPD} and older adults~\cite{Vanesa-FQAS-SRND2013}.

As one use case, personalized food recommendation is conducted over the constructed food knowledge graph FoodKG~\cite{Haussmann-FoodKG-SIP2019} with recipes, ingredients and nutrients~\cite{Chen-PFR-WSDM2021}.
When providing a recommendation, given a user query (e.g., "what is a good lunch that contains meat?") as the input, the system retrieves all recipes from FoodKG for the recommendation. Specifically, the system identifies the query type first. Then mentioned topic entity (e.g., meat) is detected from FoodKG. With the extracted entity, answers are retrieved from the knowledge graph by a KBQA model. Next, they add personalized requirements as additional constraints, such as the user's unique health conditions (e.g., allergies) and health guidelines (e.g., nutrition needs) to the raw user query for personalized food recommendation. In addition, we can obtain more accurate estimations of calories and nutrient content of the recipe to develop nutritional profiling systems via food knowledge graph-enhanced mapping between cooking recipes and structured data (food composition tables)~\cite{Azzi2020NutriSem}. Such a nutrition profiling system will further guarantee more precise dietary recommendation.

Besides, the food knowledge graph can be combined with personal knowledge graphs. A personal knowledge graph is unique for every user, and it can include personal information such as allergies, preferences, and health indexes~\cite{shirai2021applying}. With this knowledge, the reasoning of the combined food knowledge graph can be more personalized and tailored. Personal Health Knowledge Graph (PHKG) is a typical application of Semantic Web technology in comprehensive diet recommendation system~\cite{seneviratne2021personal}. This project builds a knowledge model to provide personalized dietary advice. In the project, PHKG is used to capture personal dietary behaviors such as carbohydrate intake with the extended time series summarization technique. It can also use semantic reasoners to recommend clinically relevant dietary recommendations.

\subsection{Food Supply Chain Management}

The food supply chain comprises all the stages that food products go through, from production to consumption. Nowadays with the globalization process, food is transported over longer distances before it reaches the consumers, and food supply chain thus becomes longer and more fragmented. This brings two problems, hard-achieved food traceability and more overall food waste respectively. Therefore, it is necessary to effectively manage the food supply chain to achieve reliable food traceability and control waste.

An intuitive idea is to construct a directed graph where nodes represent the status and processes of food materials. Zhang et al. first adopt the concept of Critical Tracking Events (CTEs), which is proposed by the Institute of Food Technologists (IFT) that describes the key parts of the lifecycle of a food product, like transportation and process~\cite{zhang2014guidance}. CTEs can associate with data related to the key events (like operators and devices). These CTEs can be linked and organized for tracing and tracking. However, the food supply chain system is often cross-functional and cross-region, involving data sharing problems between different company entities. For example, different food processors can have different types of data because they focus on different functions and processing, and their terms are affected by context, so the same name can indicate totally different food, like the term buttermilk: it can refer to the cultured milk drink or the milk after churning depending on where it is used~\cite{Nicholas-Review-npj2018}. Besides, food manufacturers and raw food materials suppliers may have different naming agreements due to their geography, because of which the same food and terms may have different names~\cite{Misra2020IoT}.

The food knowledge graph is a solution to model, integrate and align food data in the food supply chain management. By assigning a URI for every unique food material, the food knowledge graph can easily distinguish what exactly a term refers to under certain conditions. Besides, attributes of operations in food processing (like former step, time, status, equipment information and safety standard) once are aligned, and every participant in the food supply chain will be linked in a unified and standardized form. Based on this, more reliable traceback and related querying can be supported.

Some traceability ontologies have become a part of traceability management to support food track and trace in the food supply chain. For example, in FTTO, food processing procedures are mainly considered, and different attributes for different foods like beverages and additives are designed~\cite{TPizz-foodtrace-2013}.
Their attributes are standardized by FTTO, so different food product statuses can be linked through the processing flow and keep their naming consistency through their lifecycle. This also makes intelligent querying possible, which means the users can access all intermediate products of a certain product and their relevant information such as operating time and operator. This enables the users to trace back all security risks, such as tracing the contaminated foods, especially those that occur across borders. FTTO has been used in the global supply chain system.
MESCO further extends the FTTO to adapt the meat supply chain area. Particularly, It continues to use the traceability method and food-related concepts in FTTO, and the concept of the meat supply chain is more considered, such as the unique identification code of the animal, the place of production and the date of birth~\cite{PIZZUTI2017MESCO}.

\subsection{Food Machinery Intelligent Manufacturing}
With the continuous evolution of technologies, the innovation of IoT sensors also makes its way in food-relevant scenarios like food processing and central kitchens. Data fusion and exploration from different sensors is necessary to support further intelligent decision-making, which is of great significance for building automatic food industry production lines and consumer-oriented smart terminal equipment~\cite{iqbal2017prospects}. With the wide application of food knowledge graphs in IoT, we can build the intelligent kitchen or intelligent industry devices that can make intelligent decisions based on data from IoT sensors.

For example, in intelligent kitchens, smart refrigerators with cameras can reason with recognized food and drink items, ingredients and portion sizes, and even estimate their shelf-life for timely use in recommended recipes via the embedded food knowledge graph. Smart microwaves with cameras can recognize the food type and then automatically choose important parameters, such as heating methods and heating time via the embedded food knowledge graph.
KitchenSense is an early work about intelligent kitchens. It uses knowledge to coordinate the work of various intelligent devices in the kitchen to enhance the intelligent interaction between devices and people\cite{lee2006augmenting}.

In addition, industry robots and machinery can make more intelligent decisions from food knowledge graphs. They can access the processing status through sensors, obtain the physical properties of food materials, and perform intelligent processing control according to the knowledge from food knowledge graphs. Such an intelligent way can make information in different forms to be integrated for industrial machines. To our best knowledge, there is few published works that utilizes food knowledge graphs for food processing control. However, we notice that some food ontologies and food knowledge graphs cover the concepts involved in food processing, such as OFPE and AFEO. These established ontologies or knowledge graphs can be further explored to make different stages in food processing better controlled and more effective, resulting in more automatic and intelligent whole food processing.

\section{Future Directions of Food Knowledge Graphs}
Based on comprehensive discussions  on existing efforts, we now articulate key open challenges and future research directions for food knowledge graphs.

\subsection{Multimodal Food Knowledge Graph}
Most of existing food knowledge graphs focus on organizing  verbal knowledge extracted from text.  However, the proliferation of edge devices, such as mobile devices and IoT devices in the food industries generates large volumes of visual data, e.g., images and videos, which contain another important type of knowledge, namely visual knowledge~\cite{Perona-VOV-IEEEPro2010}.  From the narrow perspective of computer vision, visual knowledge is any information that can be useful for improving vision tasks like  recognition. Such visual knowledge includes different forms, such as  labeled examples of different categories (e.g., food categories and rich attributes) and relationships like object-object relations (e.g., Chicken is part of Kung Pow Chicken)~\cite{Chen-NEIL-ICCV2013}. Large-scale efforts, such as visual attribute learning~\cite{Ferrari2007Learning}, visual relationship detection~\cite{Cewu-VRD-ECCV2016} and scene graph generation~\cite{Xu-SGG-CVPR2017} are under way to extract a body of visual knowledge. Visual knowledge and verbal knowledge constitute multimodal knowledge. There are some initial attempts to incorporate visual information into knowledge graphs by linking images to text via hyperlinks~\cite{LiuMMKG-ESWC2019}. In this case, visual information (e.g., entity images) can only be used for visual demonstrations. Most of existing food knowledge graphs don't contain visual knowledge and thus can not support food-oriented visual search and visual illustration. It is the right time to start building a multimodal food knowledge graph, where searching, indexing, organizing and hyperlinking multimodal knowledge are necessary. Such a multimodal food knowledge graph can help food-oriented multimodal learning technologies to support many cross-modal tasks, such as cross-modal recipe-food image retrieval and generation~\cite{Salvador-LCME-TPAMI2021,Papadopoulos-Pizza-CVPR2019,WangLHM20-SAGN-ECCV2020}. The downstream applications are various, such as automatically illustrating a given recipe using semantically corresponding images, and supporting food-oriented multimodal dialogue systems. Effective multimodal information integration can also be applied to a lot of food industries scenarios, since abundant multimodal data exist in food industries, like image, video and other attribute information obtained from various types of sensors. Automatic fruit classification and grading, baking and ferment time control, and automatic food packaging can all benefit from multimodal food knowledge graphs because multimodal information can be better organized and analyzed, and these procedures can be thus optimized. However, due to different statistical properties between visual knowledge and verbal knowledge, how to reasonably and effectively build multimodal food knowledge graphs is worth further study.

\subsection{Representation and Reasoning on the Food Knowledge Graph}
The first step of using food knowledge graphs is to represent them and conduct complex reasoning on them. Numerical computing for knowledge representation and reasoning requires a continuous vector space to capture the semantics of entities and relations~\cite{Wang-KGE-TKDE2017}. While embedding-based methods have limitations on complex logical reasoning, some recently proposed methods, especially Graph Neural Networks~(GNN)~\cite{Wu-CSGNN-TNNLS2020} on knowledge graph reasoning are very promising for handling complex reasoning. The GNNs learn the representation of a target node by propagating neighbor information in an iterative manner until a stable fixed point is reached. With the help of GNNs, it is possible to extract both entity characteristics and relations from knowledge graphs, which is one essential factor for food knowledge graph-based applications, such as compound-food relation prediction~\cite{Park-FlavorGraph-SciRe2021} and food  recommendation~\cite{Wu-CSGNN-TNNLS2020}.

\subsection{Food Big Data Organization and Mining}
A huge amount of food-related information is generated globally from different sources (e.g., IoTs, online databases and social media) with various types (e.g., food components, nutrition tables, recipe text, food images and cooking videos), resulting in food big data~\cite{Tao2020Utilization}. These data are related to all the stages of the food system, such as food production, processing and consumption, and thus enable applications in the food science and industry, especially combined with AI. However, due to the large-scale of these data, how to organize and explore them well becomes a challenge.
By uniforming their names and integrating different information sources with knowledge graphs, we can better organize and utilize rich data. For example, food composition databases and food regulations can be linked with food~\cite{greenfield2003food}, and we can use them to discover connections among food and diet-induced illnesses, pushing forward relevant research on food, nutrition,  health and regulations. Furthermore, big data in the food safety domain can also help review previous food safety accidents and help develop tools to deal with hard food safety issues~\cite{CHEN201654,Hans2017Big}. If we can develop a knowledge graph to organize them, we are more likely to find the key points that lead to the accidents and resolve safety issues via reasoning on knowledge graphs. Furthermore, the big data supported food knowledge graphs can realize more precise personalized nutrition recommendations~\cite{BASHIARDES201857}. By detecting consumers' health indicators by a series of sensors, we can obtain a large amount of health data, including personal features, diet, nutrition, and behaviors. Food knowledge graphs can integrate these data which are related to the human body in an appropriate way, and the combination of food knowledge graphs and AI algorithms can realize accurate and personalized nutrition. Under appropriate circumstances, we can check health indicators again to obtain feedback, which forms a knowledge-graph-enhanced health nutrition recommendation closed-loop.

\subsection{Internet of Food (IoF)}
IoF is designed to make the data from different devices and sources interoperable and to be able to compute across the whole dataset. However, a notable limitation is the lack of integration caused by the mix of data from different sources and hardware standards. Food knowledge graphs can provide standards (food ontologies) about all food information, such as how we describe food attributes, and how  it is cooked, processed or consumed, and make all food-relevant data and information (instances) connected. Therefore, it can foster the development of IoF. However, food knowledge graphs involve complex technologies, such as knowledge aggregation, complex storing and index technologies, bringing great challenges. In addition, using knowledge graphs to integrate food data from diverse sources at a large scale is necessary~\cite{Nguyen-KGF-IF2020}, while developing scalable scientific and engineering methods to keep scale with little cost explosion is an obvious requirement for the successful application of food knowledge graphs. Once IoF is constructed based on food knowledge graphs, it enables all known food-relevant information to be accessible by machines, consumers and companies to further enable more applications in the food science and industry~\cite{Nicholas-Review-npj2018}.

\subsection{Food Knowledge Graph for Human Health}

To meet people's pursuit of better health, the essential demand is presented for better, safer and more nutritious food. To achieve this goal, building one human nutrition and health platform is necessary. The food knowledge graphs provide one opportunity to build such a platform via large-scale structured food knowledge organization. As the core of this platform, food knowledge graphs can support the tracking and monitoring of the dietary behaviors, health-relevant search and recommendations, and food-relevant studies on nutrition, diet and disease.

To achieve these goals, the food knowledge graph should satisfy some characteristics. For example, a more complete and accurate interdisciplinary food knowledge graph is one basic requirement, and joint efforts from worldwide experts in food science, nutrition, health and other relevant domains are thus needed. More challenges should also be solved. For example, there exist different culinary cultures and health beliefs in the world, which probably leads to the contradiction when adding this knowledge into the food knowledge graphs. Although existing machine learning and natural language processing methods can make food knowledge graph construction  automatically, the multi-source of food data introduces the noise inevitably. In addition, such big food knowledge graphs should support dynamic adaptation, which is more difficult to achieve from the perspective of technology.

\subsection{Food Intelligence}
Driven by the fast development of AI, there is a stringent need to push the AI frontier to the food domain. Conforming to this trend, food computing~\cite{Min2019A} has  received tremendous amounts of interest for its multifarious applications in health, culture and medicine. It acquires and analyzes multi-source multimodal food data for food-oriented various tasks via computational approaches. The nexus between food computing and AI gives birth to the novel paradigm of food intelligence. Food knowledge graphs can enhance already popular techniques of computer vision and natural language processing, such as image recognition~\cite{Marino-UKGIC-CVPR2019,KNEZ2020Food}, object detection~\cite{FangKLTC17} and QA~\cite{zhang2018variational}, and thus can aid food computing tasks. We can also make decisions and reasoning on food knowledge graphs~\cite{KRKG-Xiaojun-ESA2020} in combination with advanced AI technologies for many intelligent services in various fields, such as smart kitchen~\cite{Krieg-CookingAssistance2015}. Therefore, food knowledge graphs will play important roles in realizing food intelligence, which will benefit various studies and applications in the food science and industry.

\section{Conclusions}
In this review, we summarize  food knowledge graphs from their development, applications, and future directions in the food science and industry. Our comprehensive review of current research on food knowledge graphs shows that food knowledge graphs have enabled various food-oriented applications for the capability of  knowledge graphs in effective food data organization, representation and reasoning. Future directions for food knowledge graphs show their great potential in solving food-relevant key problems  in food industries and daily diet scenarios. Although there are still challenges from  multimodal food data and complex computational technologies, we have seen the considerable application prospects shown by food knowledge graphs in the food domain. This is also the purpose of this review, which  encourages researchers and engineers in this field to put knowledge graphs into practice for the benefits of food science and industry. 


\section*{Acknowledgments}
The authors gratefully acknowledge the financial support from the National Natural Science Foundation of China under Grant 61972378, U1936203, U19B2040.

\section*{Author contributions}
Weiqing Min: Conceptualization-Equal, Data curation-Equal, Investigation-Lead, Writing – original draft-Lead, Writing – review \& editing-Equal. Chunlin Liu: Data curation-Equal, Investigation-Equal, Writing – original draft-Equal, Writing – review \& editing-Equal. Leyi Xu: Investigation-Equal, Visualization-Equal, Writing – review \& editing-Equal. Shuqiang Jiang: Conceptualization-Lead, Formal analysis-Equal, Funding acquisition-Lead, Writing – original draft-Equal, Writing – review \& editing-Equal.

\section*{Declaration of interests}
No conflict of interests declared.

\end{spacing}

\bibliographystyle{unsrt} 
\bibliography{FKG_NF}  

\begin{thebibliography}{100}

\bibitem{FFS-Report2020}
{Global Panel on Agriculture and Food Systems for Nutrition}.
\newblock Future food systems: For people, our planet, and prosperity.
\newblock Technical report, Global Panel on Agriculture and Food Systems for
  Nutrition, 2020.

\bibitem{Raj2020Comprehensive}
G.~V. S.~Bhagya Raj and Kshirod~K. Dash.
\newblock Comprehensive study on applications of artificial neural network in
  food process modeling.
\newblock {\em Critical Reviews in Food Science and Nutrition}, pages 1--28,
  2020.

\bibitem{Qing2019Recent}
Qing Sun, Min Zhang, and Arun~S. Mujumdar.
\newblock Recent developments of artificial intelligence in drying of fresh
  food: A review.
\newblock {\em Critical Reviews in Food Science and Nutrition},
  59(14):2258--2275, 2019.

\bibitem{Amani2020Current}
Hanieh Amani, Katalin Kerti~Badakné, and Amin Mousavi~Khaneghah.
\newblock Current progress in the utilization of smartphone-based imaging for
  quality assessment of food products: A review.
\newblock {\em Critical Reviews in Food Science and Nutrition}, pages 1--13, 12
  2020.

\bibitem{KNEZ2020Food}
Simon Knez and Luka Šajn.
\newblock Food object recognition using a mobile device: Evaluation of
  currently implemented systems.
\newblock {\em Trends in Food Science \& Technology}, 99:460--471, 2020.

\bibitem{Marcus-VMVL-Patterns2020}
Marcus Klasson, Cheng Zhang, and Hedvig Kjellstr{\"{o}}m.
\newblock Using variational multi-view learning for classification of grocery
  items.
\newblock {\em Patterns}, 1(8):100143, 2020.

\bibitem{LIU2021Efficient}
Yao Liu, Hongbin Pu, and Da-Wen Sun.
\newblock Efficient extraction of deep image features using convolutional
  neural network ({CNN}) for applications in detecting and analysing complex
  food matrices.
\newblock {\em Trends in Food Science \& Technology}, 113:193--204, 2021.

\bibitem{Liang-AIS-CRFSN2020}
Ning Liang, Sashuang Sun, Chu Zhang, Yong He, and Zhengjun Qiu.
\newblock Advances in infrared spectroscopy combined with artificial neural
  network for the authentication and traceability of food.
\newblock {\em Critical Reviews in Food Science and Nutrition}, pages 1--22,
  2020.

\bibitem{thames2021nutrition5k}
Quin Thames, Arjun Karpur, Wade Norris, Fangting Xia, Liviu Panait, Tobias
  Weyand, and Jack Sim.
\newblock Nutrition5k: Towards automatic nutritional understanding of generic
  food.
\newblock In {\em Proceedings of the IEEE/CVF Conference on Computer Vision and
  Pattern Recognition}, pages 8903--8911, 2021.

\bibitem{Larissa2021App}
Larissa~Oliveira Chaves, Ana Luiza~Gomes Domingos, Daniel~Louzada Fernandes,
  Fabio~Ribeiro Cerqueira, Rodrigo Siqueira-Batista, and Josefina Bressan.
\newblock Applicability of machine learning techniques in food intake
  assessment: A systematic review.
\newblock {\em Critical Reviews in Food Science and Nutrition}, pages 1--18,
  2021.

\bibitem{Misra2020IoT}
N.~N. Misra, Yash Dixit, Ahmad Al-Mallahi, Manreet~Singh Bhullar, Rohit
  Upadhyay, and Alex Martynenko.
\newblock {IoT}, big data and artificial intelligence in agriculture and food
  industry.
\newblock {\em IEEE Internet of Things Journal}, pages 1--1, 2020.

\bibitem{Qian2020Traceability}
Jianping Qian, Bingye Dai, Baogang Wang, Yan Zha, and Qian Song.
\newblock Traceability in food processing: Problems, methods, and performance
  evaluations—a review.
\newblock {\em Critical Reviews in Food Science and Nutrition}, pages 1--14,
  2020.

\bibitem{Nicholas-Review-npj2018}
Nicholas~M. Holden, Eoin~P. White, Matthew.~C. Lange, and Thomas~L. Oldfield.
\newblock Review of the sustainability of food systems and transition using the
  {Internet of Food}.
\newblock {\em npj Science of Food}, 2(1):18, 2018.

\bibitem{Igor-OKM-KIS2004}
Igor Jurisica, John Mylopoulos, and Eric Yu.
\newblock Ontologies for knowledge management: An information systems
  perspective.
\newblock {\em Knowledge and Information Systems}, 6:380--401, 2004.

\bibitem{Dooley-FoodOn-npj2019}
Damion~M. Dooley, Emma~J. Griffiths, Gurinder~S. Gosal, Pier~L. Buttigieg,
  Robert Hoehndorf, Matthew~C. Lange, Lynn~M. Schriml, Fiona S.~L. Brinkman,
  and William W.~L. Hsiao.
\newblock {FoodOn}: A harmonized food ontology to increase global food
  traceability, quality control and data integration.
\newblock {\em npj Science of Food}, 2, 2018.

\bibitem{Eftimov-ISO-FOOD-FC2019}
Tome Eftimov, Gordana Ispirova, Doris Potočnik, Nives Ogrinc, and
  Barbara~Koroušić Seljak.
\newblock {ISO-FOOD} ontology: A formal representation of the knowledge within
  the domain of isotopes for food science.
\newblock {\em Food Chemistry}, 277:382--390, 2019.

\bibitem{Elizabeth-OCP-Patterns2020}
Elizabeth Arnaud, Marie-Ang{\'e}lique Laporte, Soonho Kim, C{\'e}line Aubert,
  Sabina Leonelli, Berta Miro, Laurel Cooper, Pankaj Jaiswal, Gideon Kruseman,
  Rosemary Shrestha, Pier~Luigi Buttigieg, Christopher~J. Mungall, Julian
  Pietragalla, Afolabi Agbona, Jacqueline Muliro, Jeffrey Detras, Vilma Hualla,
  Abhishek Rathore, Roma~Rani Das, Ibnou Dieng, Guillaume Bauchet, Naama Menda,
  Cyril Pommier, Felix Shaw, David Lyon, Leroy Mwanzia, Henry Juarez, Enrico
  Bonaiuti, Brian Chiputwa, Olatunbosun Obileye, Sandrine Auzoux,
  Esther~Dzal{\'e} Yeumo, Lukas~A. Mueller, Kevin Silverstein, Alexandra
  Lafargue, Erick Antezana, Medha Devare, and Brian King.
\newblock The ontologies community of practice: A {CGIAR} initiative for big
  data in agrifood systems.
\newblock {\em Patterns}, 1(7):100105, 2020.

\bibitem{paulheim2017knowledge}
Heiko Paulheim.
\newblock Knowledge graph refinement: A survey of approaches and evaluation
  methods.
\newblock {\em Semantic Web}, 8(3):489--508, 2017.

\bibitem{Haussmann-FoodKG-SIP2019}
Steven Haussmann, Oshani Seneviratne, Yu~Chen, Yarden Ne'eman, James Codella,
  Ching-Hua Chen, Deborah~L. McGuinness, and Mohammed~J. Zaki.
\newblock {FoodKG}: A semantics-driven knowledge graph for food recommendation.
\newblock In {\em The Semantic Web -- International Semantic Web Conference},
  pages 146--162, 2019.

\bibitem{Wang-KGE-TKDE2017}
Q.~{Wang}, Z.~{Mao}, B.~{Wang}, and L.~{Guo}.
\newblock Knowledge graph embedding: A survey of approaches and applications.
\newblock {\em IEEE Transactions on Knowledge and Data Engineering},
  29(12):2724--2743, 2017.

\bibitem{KRKG-Xiaojun-ESA2020}
Xiaojun Chen, Shengbin Jia, and Yang Xiang.
\newblock A review: Knowledge reasoning over knowledge graph.
\newblock {\em Expert Systems with Applications}, 141:112948, 2020.

\bibitem{Shaoxiong-KG-arXiv2020}
Shaoxiong Ji, Shirui Pan, Erik Cambria, Pekka Marttinen, and Philip~S. Yu.
\newblock A survey on knowledge graphs: Representation, acquisition, and
  applications.
\newblock {\em IEEE Transactions on Neural Networks and Learning Systems},
  pages 1--21, 2021.

\bibitem{Hogan-KG-CSUR2021}
Aidan Hogan, Eva Blomqvist, Michael Cochez, Claudia d'Amato, Gerard de~Melo,
  Claudio Guti{\'{e}}rrez, Sabrina Kirrane, Jos{\'{e}} Emilio~Labra Gayo,
  Roberto Navigli, Sebastian Neumaier, Axel{-}Cyrille~Ngonga Ngomo, Axel
  Polleres, Sabbir~M. Rashid, Anisa Rula, Lukas Schmelzeisen, Juan~F. Sequeda,
  Steffen Staab, and Antoine Zimmermann.
\newblock Knowledge graphs.
\newblock {\em {ACM} Comput. Surv.}, 54(4):71:1--71:37, 2021.

\bibitem{collins1969retrieval}
Allan~M Collins and M~Ross Quillian.
\newblock Retrieval time from semantic memory.
\newblock {\em Journal of verbal learning and verbal behavior}, 8(2):240--247,
  1969.

\bibitem{berners1998semantic}
Tim Berners-Lee.
\newblock Semantic web road map, 1998.

\bibitem{berners2006tabulator}
Tim Berners-Lee, Yuhsin Chen, Lydia Chilton, Dan Connolly, Ruth Dhanaraj, James
  Hollenbach, Adam Lerer, and David Sheets.
\newblock Tabulator: Exploring and analyzing linked data on the semantic web.
\newblock In {\em Proceedings of the international semantic web user
  interaction workshop}, volume 2006, page 159, 2006.

\bibitem{Gertjan-CSIO-AIM1995}
Gertjan van Heijst, Sabina Falasconi, Ameen Abu-Hanna, Guus Schreiber, and
  Mario Stefanelli.
\newblock A case study in ontology library construction.
\newblock {\em Artificial Intelligence in Medicine}, 7(3):227--255, 1995.

\bibitem{google2012kg}
Amit Singhal.
\newblock Introducing the knowledge graph: Things, not strings, 2012.

\bibitem{Nickel-RMLKG-IEEEPro2016}
M.~{Nickel}, K.~{Murphy}, V.~{Tresp}, and E.~{Gabrilovich}.
\newblock A review of relational machine learning for knowledge graphs.
\newblock {\em Proceedings of the {IEEE}}, 104(1):11--33, 2016.

\bibitem{Lenat-1995-CYC}
Douglas~B. Lenat.
\newblock {CYC}: A large-scale investment in knowledge infrastructure.
\newblock {\em Communications of the {ACM}}, 38(11):33--38, November 1995.

\bibitem{Denny2014wikidata}
Denny Vrandečić and Markus Krötzsch.
\newblock Wikidata: A free collaborative knowledge base.
\newblock {\em Communications of the {ACM}}, 57:78--85, 2014.

\bibitem{Lehmann2014DBpedia}
Jens Lehmann, Robert Isele, Max Jakob, Anja Jentzsch, Dimitris Kontokostas,
  Pablo Mendes, Sebastian Hellmann, Mohamed Morsey, Patrick Van~Kleef, Sören
  Auer, and Christian Bizer.
\newblock {DBpedia} - a large-scale, multilingual knowledge base extracted from
  {Wikipedia}.
\newblock {\em Semantic Web}, 6(2):167--195, 01 2014.

\bibitem{Luna-KV-KDD2014}
Xin~Luna Dong, Evgeniy Gabrilovich, Geremy Heitz, Wilko Horn, Ni~Lao, Kevin
  Murphy, Thomas Strohmann, Shaohua Sun, and Wei Zhang.
\newblock {Knowledge Vault}: A web-scale approach to probabilistic knowledge
  fusion.
\newblock In {\em Proceedings of {ACM} {SIGKDD} International Conference on
  Knowledge Discovery and Data Mining}, pages 601--610, 2014.

\bibitem{Han-CEL-SIGIR2011}
Xianpei Han, Le~Sun, and Jun Zhao.
\newblock Collective entity linking in web text: A graph-based method.
\newblock In {\em Proceedings of International {ACM} {SIGIR} Conference on
  Research and Development in Information Retrieval}, pages 765--774, 2011.

\bibitem{zhang2020transrhs}
Fuxiang Zhang, Xin Wang, Zhao Li, and Jianxin Li.
\newblock {TransRHS}: A representation learning method for knowledge graphs
  with relation hierarchical structure.
\newblock In {\em Proceedings of International Joint Conference on Artificial
  Intelligence}, pages 2987--2993, 2020.

\bibitem{nayyeri20215}
Mojtaba Nayyeri, Sahar Vahdati, Can Aykul, and Jens Lehmann.
\newblock 5* knowledge graph embeddings with projective transformations.
\newblock In {\em Proceedings of the {AAAI} Conference on Artificial
  Intelligence}, volume~35, pages 9064--9072, 2021.

\bibitem{cenikj2021foodchem}
Gjorgjina Cenikj, Barbara~Korou{\v{s}}i{\'c} Seljak, and Tome Eftimov.
\newblock {FoodChem}: A food-chemical relation extraction model.
\newblock {\em arXiv preprint arXiv:2110.02019}, 2021.

\bibitem{Tay-MTNN-CIKM2017}
Yi~Tay, Luu~Anh Tuan, Minh~C. Phan, and Siu~Cheung Hui.
\newblock Multi-task neural network for non-discrete attribute prediction in
  knowledge graphs.
\newblock In {\em Proceedings of {ACM} on Conference on Information and
  Knowledge Management}, pages 1029--1038, 2017.

\bibitem{Hirschman-NLQA-NLE2001}
L.~Hirschman and R.~Gaizauskas.
\newblock Natural language question answering: The view from here.
\newblock {\em Natural Language Engineering}, 7(4):275--300, 2001.

\bibitem{Nilesh-NNA-CoRR2019}
Nilesh Chakraborty, Denis Lukovnikov, Gaurav Maheshwari, Priyansh Trivedi, Jens
  Lehmann, and Asja Fischer.
\newblock Introduction to neural network based approaches for question
  answering over knowledge graphs.
\newblock {\em arXiv preprint arXiv:1907.09361}, 2019.

\bibitem{Oramas-SMR-TIST2017}
Sergio Oramas, Vito~Claudio Ostuni, Tommaso~Di Noia, Xavier Serra, and
  Eugenio~Di Sciascio.
\newblock Sound and music recommendation with knowledge graphs.
\newblock {\em {ACM} Transactions on Intelligent Systems and Technology},
  8(2):1--21, 2016.

\bibitem{Hongwei-RippleNet-CIKM2018}
Hongwei Wang, Fuzheng Zhang, Jialin Wang, Miao Zhao, Wenjie Li, Xing Xie, and
  Minyi Guo.
\newblock {RippleNet}: Propagating user preferences on the knowledge graph for
  recommender systems.
\newblock In {\em Proceedings of {ACM} International Conference on Information
  and Knowledge Management}, pages 417--426, 2018.

\bibitem{Boulos2015Towards}
Maged N~Kamel Boulos, Abdulslam Yassine, Shervin Shirmohammadi, Chakkrit~Snae
  Namahoot, and Michael Brückner.
\newblock Towards an {“Internet of Food”}: Food ontologies for the
  {Internet of Things}.
\newblock {\em Future Internet}, 7(4):372--392, 2015.

\bibitem{Cantais-FODC-SWeb2005}
Jaime Cantais, David Dominguez, Valeria Gigante, Loredana Laera, and Valentina
  Tamma.
\newblock An example of food ontology for diabetes control.
\newblock In {\em Proceedings of the International Semantic Web Conference 2005
  workshop on Ontology Patterns for the Semantic Web}, 2005.

\bibitem{Batista-OCcooking-2006}
Fernando Batista, J.~Paulo, Nuno Mamede, H.~Vaz, and Ricardo Ribeiro.
\newblock Ontology construction: Cooking domain.
\newblock {\em Artificial Intelligence: Methodology, Systems, and
  Applications}, 4183:213--221, 01 2006.

\bibitem{Snae-FOODS-ICDET2008}
C.~{Snae} and M.~{Bruckner}.
\newblock {FOODS}: A food-oriented ontology-driven system.
\newblock In {\em IEEE International Conference on Digital Ecosystems and
  Technologies}, pages 168--176, 2008.

\bibitem{Caracciolo-TMAPLD-MSR2011}
Caterina Caracciolo, Ahsan Morshed, Armando Stellato, Gudrun Johannsen, Yves
  Jaques, and Johannes Keizer.
\newblock Thesaurus maintenance, alignment and publication as linked data: The
  {AGROOVOC} use case.
\newblock In {\em Metadata and Semantic Research}, pages 489--499, 2011.

\bibitem{TPizz-foodtrace-2013}
T.~{Pizzuti} and G.~{Mirabelli}.
\newblock {FTTO}: An example of food ontology for traceability purpose.
\newblock In {\em IEEE International Conference on Intelligent Data Acquisition
  and Advanced Computing Systems}, volume~01, pages 281--286, Sep. 2013.

\bibitem{Cordier-Taaable-2014}
Am{\'e}lie Cordier, Valmi Dufour-Lussier, Jean Lieber, Emmanuel Nauer, Fadi
  Badra, Julien Cojan, Emmanuelle Gaillard, Laura Infante-Blanco, Pascal Molli,
  Amedeo Napoli, and Hala Skaf-Molli.
\newblock Taaable: A case-based system for personalized cooking.
\newblock {\em Studies in Computational Intelligence}, 494:121--162, 01 2014.

\bibitem{Karim-Travelers-2015}
Shakir Karim, Umair Shaikh, Quratulain Rajput, and Zaheeruddin Asif.
\newblock Ontology-based personalized dietary recommendation for travelers.
\newblock In {\em Southern association for information systems conference},
  2015.

\bibitem{Peroni-FOod-2016}
Silvio Peroni, Giorgia Lodi, Luigi Asprino, Aldo Gangemi, and Valentina
  Presutti.
\newblock {FOOD: FOod in Open Data}.
\newblock In {\em The Semantic Web-International Semantic Web Conference},
  pages 168--176, 2016.

\bibitem{Aelik-FoodWiki-SWJ2015}
Duygu {\c{C}}elik~Ertu{\u{g}}rul.
\newblock {FoodWiki}: Ontology-driven mobile safe food consumption system.
\newblock {\em The Scientific world journal}, 2015:475410, 2015.

\bibitem{Maxim-FOODpedia-ESWC2015}
Maxim Kolchin, Alexander Chistyakov, Maxim Lapaev, and Rezeda Khaydarova.
\newblock {FOODpedia}: Russian food products as a linked data dataset.
\newblock In {\em The Semantic Web: Extended Semantic Web Conference 2015
  Satellite Events}, pages 87--90, 2015.

\bibitem{Ibanescu2016PO2}
Liliana Ibanescu, Juliette Dibie, St{\'e}phane Dervaux, Elisabeth Guichard, and
  Joe Raad.
\newblock $po^2$- a process and observation ontology in food science.
  application to dairy gels.
\newblock In Emmanouel Garoufallou, Imma Subirats~Coll, Armando Stellato, and
  Jane Greenberg, editors, {\em Metadata and Semantics Research}, pages
  155--165, 2016.

\bibitem{Eftimov2018315}
Tome Eftimov, Gordana Ispirova, Peter Korošec, and Barbara Seljak.
\newblock The {RICHFIELDS} framework for semantic interoperability of food
  information across heterogenous information systems.
\newblock In {\em Proceedings of International Joint Conference on Knowledge
  Discovery, Knowledge Engineering and Knowledge Management}, volume~1, pages
  315--322, 09 2018.

\bibitem{PIZZUTI2017MESCO}
Teresa Pizzuti, Giovanni Mirabelli, Giovanni Grasso, and Giulia Paldino.
\newblock {MESCO} (meat supply chain ontology): An ontology for supporting
  traceability in the meat supply chain.
\newblock {\em Food Control}, 72:123--133, 2017.

\bibitem{Mauro-HeLiS-ISWC2018}
Mauro Dragoni, Tania Bailoni, Rosa Maimone, and Claudio Eccher.
\newblock {HeLiS}: An ontology for supporting healthy lifestyles.
\newblock In {\em The Semantic Web -- International Semantic Web Conference
  2018}, pages 53--69, 2018.

\bibitem{Vitali-ONS-GeNutri2018}
Francesco Vitali, Rosario Lombardo, Damariz Rivero, Fulvio Mattivi, Pietro
  Franceschi, Alessandra Bordoni, Alessia Trimigno, Francesco Capozzi, Giovanni
  Felici, Francesco Taglino, Franco Miglietta, Nathalie De~Cock, Carl Lachat,
  Bernard De~Baets, Guy De~Tr{\'e}, Mariona Pinart, Katharina Nimptsch, Tobias
  Pischon, Jildau Bouwman, Duccio Cavalieri, and {the {ENPADASI} consortium}.
\newblock {ONS}: An ontology for a standardized description of interventions
  and observational studies in nutrition.
\newblock {\em Genes \& Nutrition}, 13(1):12, 2018.

\bibitem{Qin2019FSKG}
Li~Qin, Zhigang Hao, and Liang Zhao.
\newblock Food safety knowledge graph and question answering system.
\newblock In {\em Proceedings of International Conference on Information
  Technology: IoT and Smart City}, pages 559--564, 2019.

\bibitem{Caste2020FOBI}
Pol Castellano-Escuder, Raúl González-Domínguez, David~S Wishart, Cristina
  Andrés-Lacueva, and Alex Sánchez-Pla.
\newblock {{FOBI}: An Ontology to Represent Food Intake Data and Associate it
  with Metabolomic Data}.
\newblock {\em Database}, 2020, 06 2020.

\bibitem{Ameri2020Enabling}
Farhad Ameri, Evan Wallace, and Reid Yoder.
\newblock {Enabling Traceability in Agri-Food Supply Chains Using an
  Ontological Approach}.
\newblock In {\em International Design Engineering Technical Conferences and
  Computers and Information in Engineering Conference}, volume 83983, page
  V009T09A053, 08 2020.

\bibitem{sherimon2021modeling}
Vinu Sherimon, PC~Sherimon, Alaa Ismaeel, Winny Varkey, and B~Naveen.
\newblock Modeling of seafood domain using ontology.
\newblock {\em International Journal of Open Information Technologies},
  9(2):65--69, 2021.

\bibitem{Padhiar2021SemanticMF}
Ishita Padhiar, Oshani~Wasana Seneviratne, Shruthi Chari, Daniel Gruen, and
  Deborah~L. McGuinness.
\newblock Semantic modeling for food recommendation explanations.
\newblock pages 13--19, 2021.

\bibitem{amith2021ontology}
Muhammad Amith, Chidinma Onye, Tracey Ledoux, Grace Xiong, and Cui Tao.
\newblock The ontology of fast food facts: conceptualization of nutritional
  fast food data for consumers and semantic web applications.
\newblock {\em BMC medical informatics and decision making}, 21(7):1--16, 2021.

\bibitem{Min-YAWYE-TMM2018}
Weiqing Min, Bing-Kun Bao, Shuhuan Mei, Yaohui Zhu, Yong Rui, and Shuqiang
  Jiang.
\newblock You are what you eat: Exploring rich recipe information for
  cross-region food analysis.
\newblock {\em {IEEE} Transactions on Multimedia}, 20(4):950--964, 2018.

\bibitem{Sajadmanesh-KC-arXiv2016}
Sina Sajadmanesh, Sina Jafarzadeh, Seyed~Ali Ossia, Hamid~R Rabiee, Hamed
  Haddadi, Yelena Mejova, Mirco Musolesi, Emiliano De~Cristofaro, and Gianluca
  Stringhini.
\newblock Kissing cuisines: Exploring worldwide culinary habits on the web.
\newblock In {\em Proceedings of International Conference on World Wide Web
  Companion}, pages 1013--1021, 2017.

\bibitem{CelikErtugru-FoodWiki-SWJ2015}
Duygu {\c{C}}elik~Ertu{\u{g}}rul.
\newblock {FoodWiki}: {A} mobile app examines side effects of food additives
  via semantic web.
\newblock {\em Journal of Medical Systems}, 40(2):41, Nov 2015.

\bibitem{Teresa-FTTO-JFE2014}
Teresa Pizzuti, Giovanni Mirabelli, Miguel~Angel Sanz-Bobi, and Fernando
  Goméz-Gonzaléz.
\newblock Food track \& trace ontology for helping the food traceability
  control.
\newblock {\em Journal of Food Engineering}, 120:17--30, 2014.

\bibitem{muljarto2014ontology}
Aunur-Rofiq Muljarto, Jean-Michel Salmon, Pascal Neveu, Brigitte Charnomordic,
  and Patrice Buche.
\newblock Ontology-based model for food transformation processes-application to
  winemaking.
\newblock In {\em Research Conference on Metadata and Semantics Research},
  pages 329--343, 2014.

\bibitem{popovski2019foodontomap}
Gorjan Popovski, Barbara Korousic-Seljak, and Tome Eftimov.
\newblock Foodontomap: Linking food concepts across different food ontologies.
\newblock In {\em Proceedings of the 11th International Conference on Knowledge
  Engineering and Ontology Development}, pages 195--202, 2019.

\bibitem{ChiKnowledge}
Yang Chi, Congcong Yu, Xiaohui Qi, and Hao Xu.
\newblock Knowledge management in healthcare sustainability: A smart healthy
  diet assistant in traditional {Chinese} medicine culture.
\newblock {\em Sustainability}, 10(11), 2018.

\bibitem{Zulaika-EPCARFS-Proceedings2018}
Unai Zulaika, Asier Guti{\'{e}}rrez, and Diego L{\'{o}}pez{-}de{-}Ipi{\~{n}}a.
\newblock Enhancing profile and context aware relevant food search through
  knowledge graphs.
\newblock In {\em International Conference on Ubiquitous Computing and Ambient
  Intelligence}, volume~2, page 1228, 2018.

\bibitem{Huang2019Towards}
Lan Huang, Congcong Yu, Yang Chi, Xiaohui Qi, and Hao Xu.
\newblock Towards smart healthcare management based on knowledge graph
  technology.
\newblock In {\em Proceedings of the International Conference on Software and
  Computer Applications}, page 330–337, 2019.

\bibitem{Yuanzhe-AgriKG-DSAA2019}
Yuanzhe Chen, Jun Kuang, Dawei Cheng, Jianbin Zheng, Ming Gao, and Aoying Zhou.
\newblock {AgriKG}: An agricultural knowledge graph and its applications.
\newblock In {\em Database Systems for Advanced Applications}, pages 533--537,
  2019.

\bibitem{Steven-FoodKG-ISWC2019-1}
Haussmann Steven, Chen Yu, Seneviratne Oshani, Rastogi Nidhi, Codella James,
  Chen Ching-Hua, McGuinness Deborah, and J.~Zaki Mohammed.
\newblock {FoodKG} enabled {Q\&A} application.
\newblock In {\em Proceedings of International Semantic Web Conference}, 2019.

\bibitem{milanlouei2020systematic}
Soodabeh Milanlouei, Giulia Menichetti, Yanping Li, Joseph Loscalzo, Walter~C
  Willett, and Albert-L{\'a}szl{\'o} Barab{\'a}si.
\newblock A systematic comprehensive longitudinal evaluation of dietary factors
  associated with acute myocardial infarction and fatal coronary heart disease.
\newblock {\em Nature communications}, 11(1):1--14, 2020.

\bibitem{Qin2020Question}
Li~Qin, Zhigang Hao, and LiPing Yang.
\newblock Question answering system based on food spot-check knowledge graph.
\newblock In {\em Proceedings of International Conference on Computing and Data
  Engineering}, ICCDE 2020, page 168–172, 2020.

\bibitem{Rostami2021WFAP}
Ali Rostami, Zhouhang Xie, Akihisa Ishino, Yoko Yamakata, Kiyoharu Aizawa, and
  Ramesh Jain.
\newblock {World Food Atlas Project}.
\newblock In {\em Proceedings of International Workshop on Multimedia for
  Cooking and Eating Activities}, pages 33--36, 2021.

\bibitem{LEI2021115708}
Zhenfeng Lei, Anwar {Ul Haq}, Adnan Zeb, Md~Suzauddola, and Defu Zhang.
\newblock Is the suggested food your desired?: Multi-modal recipe
  recommendation with demand-based knowledge graph.
\newblock {\em Expert Systems with Applications}, 186:115708, 2021.

\bibitem{Salvador-LCME-TPAMI2021}
Javier Mar{\'{\i}}n, Aritro Biswas, Ferda Ofli, Nicholas Hynes, Amaia Salvador,
  Yusuf Aytar, Ingmar Weber, and Antonio Torralba.
\newblock {Recipe1M+}: {A} dataset for learning cross-modal embeddings for
  cooking recipes and food images.
\newblock {\em {IEEE} Transactions on Pattern Analysis and Machine
  Intelligence}, 43(1):187--203, 2021.

\bibitem{shirai2020semantics}
Sola Shirai, Oshani Seneviratne, Minor Gordon, Ching-Hua Chen, Deborah~L
  McGuinness, Nidhi Rastogi, et~al.
\newblock Semantics-driven ingredient substitution in the {FoodKG}.
\newblock In {\em The Semantic Web -- International Semantic Web Conference},
  volume 2721, pages 242--247, 2020.

\bibitem{yang2005chinese}
Yuexin Yang, GY~Wang, and XC~Pan.
\newblock Chinese food composition table 2004, 2005.

\bibitem{Veron2020A}
Mathilde Veron, Anselmo Pe{\~{n}}as, Guillermo Echegoyen, Somnath Banerjee,
  Sahar Ghannay, and Sophie Rosset.
\newblock A cooking knowledge graph and benchmark for question answering
  evaluation in lifelong learning scenarios.
\newblock In {\em Natural Language Processing and Information Systems}, pages
  94--101, 2020.

\bibitem{Natasha-InSKG-ACMCommun2019}
Natasha Noy, Yuqing Gao, Anshu Jain, Anant Narayanan, Alan Patterson, and Jamie
  Taylor.
\newblock Industry-scale knowledge graphs: Lessons and challenges.
\newblock {\em Communications of the {ACM}}, 62(8):36--43, 2019.

\bibitem{uber2018kg}
Hamad Ferras, Liu Isaac, and Zhang Xianxing.
\newblock Food discovery with {Uber Eats}: Building a query understanding
  engine, 6 2018.

\bibitem{HELMY20151071}
Tarek Helmy, Ahmed Al-Nazer, Saeed Al-Bukhitan, and Ali Iqbal.
\newblock Health, food and user's profile ontologies for personalized
  information retrieval.
\newblock {\em Procedia Computer Science}, 52:1071--1076, 2015.

\bibitem{Virglio-OBP-2005}
João Graça, Marcio~Duarte Albasini~Mourao, Orlando Anunciação, Pedro
  Monteiro, H.~Sofia Pinto, and Virgílio Loureiro.
\newblock Ontology building process: The wine domain.
\newblock {\em Conference of the European Federation for Information Technology
  in Agriculture, Food and Environment}, pages 1138--1145, 2005.

\bibitem{Popovski-FoodBase-Database2019}
Gorjan Popovski, Barbara~Koroušić Seljak, and Tome Eftimov.
\newblock {FoodBase} corpus: A new resource of annotated food entities.
\newblock {\em Database}, 2019, 2019.

\bibitem{Popovski-FNE-Access2020}
G.~{Popovski}, B.~K. {Seljak}, and T.~{Eftimov}.
\newblock A survey of named-entity recognition methods for food information
  extraction.
\newblock {\em IEEE Access}, 8:31586--31594, 2020.

\bibitem{DietHub-TFST-2021}
Matej Petković, Gorjan Popovski, Barbara~Koroušić Seljak, Dragi Kocev, and
  Tome Eftimov.
\newblock {DietHub}: Dietary habits analysis through understanding the content
  of recipes.
\newblock {\em Trends in Food Science \& Technology}, 107:183--194, 2021.

\bibitem{cenikj2021saffron}
Gjorgjina Cenikj, Tome Eftimov, and Barbara~Korou{\v{s}}i{\'c} Seljak.
\newblock Saffron: transfer learning for food-disease relation extraction.
\newblock In {\em Proceedings of the 20th Workshop on Biomedical Language
  Processing}, pages 30--40, 2021.

\bibitem{Shirai-IISKGF-FAI2021}
Sola~S. Shirai, Oshani Seneviratne, Minor~E. Gordon, Ching-Hua Chen, and
  Deborah~L. McGuinness.
\newblock Identifying ingredient substitutions using a knowledge graph of food.
\newblock {\em Frontiers in Artificial Intelligence}, 3, 2021.

\bibitem{Pinel2015}
Florian Pinel, Lav~R. Varshney, and Debarun Bhattacharjya.
\newblock {\em A Culinary Computational Creativity System}, pages 327--346.
\newblock Atlantis Press, Paris, 2015.

\bibitem{mikolov2013efficient}
Tomas Mikolov, Kai Chen, Greg Corrado, and Jeffrey Dean.
\newblock Efficient estimation of word representations in vector space.
\newblock {\em arXiv preprint arXiv:1301.3781}, 2013.

\bibitem{Ahn-FNFP-SciRe2011}
Yong-Yeol Ahn, Sebastian~E Ahnert, James~P Bagrow, and Albert-L{\'a}szl{\'o}
  Barab{\'a}si.
\newblock Flavor network and the principles of food pairing.
\newblock {\em Scientific reports}, 1, 2011.

\bibitem{Davis2016Comparative}
Allan~Peter Davis, Cynthia~J. Grondin, Robin~J. Johnson, Daniela Sciaky,
  Benjamin~L. King, Roy McMorran, Jolene Wiegers, Thomas~C. Wiegers, and
  Carolyn~J. Mattingly.
\newblock {The Comparative Toxicogenomics Database: Update 2017}.
\newblock {\em Nucleic Acids Research}, 45(D1):D972--D978, 09 2016.

\bibitem{Wishart2006Drugbank}
David~S. Wishart, Craig Knox, An~Chi Guo, Savita Shrivastava, Murtaza
  Hassanali, Paul Stothard, Zhan Chang, and Jennifer Woolsey.
\newblock {DrugBank}: A comprehensive resource for in silico drug discovery and
  exploration.
\newblock {\em Nucleic Acids Research}, 34(suppl\_1):D668--D672, 01 2006.

\bibitem{Riyanka-CookingQA-SIP2017}
Riyanka Manna, Partha Pakray, Somnath Banerjee, Dipankar Das, and Alexander
  Gelbukh.
\newblock {CookingQA}: A question answering system based on cooking ontology.
\newblock In {\em Advances in Computational Intelligence}, pages 67--78, 2017.

\bibitem{Semih-RecipeQA-EMNLP2018}
Yagcioglu Semih, Erdem Aykut, Erdem Erkut, and Ikizler-Cinbis Nazli.
\newblock {RecipeQA}: A challenge dataset for multimodal comprehension of
  cooking recipes.
\newblock In {\em Proceedings of Conference on Empirical Methods in Natural
  Language Processing}, pages 1358--1368, 2018.

\bibitem{woodside2013fruit}
Jayne~V Woodside, Ian~S Young, and Michelle~C McKinley.
\newblock Fruit and vegetable intake and risk of cardiovascular disease.
\newblock {\em Proceedings of the Nutrition Society}, 72(4):399--406, 2013.

\bibitem{Ashkan2019Health}
Ashkan Afshin, Patrick~John Sur, Kairsten~A. Fay, Leslie Cornaby, Giannina
  Ferrara, Joseph~S Salama, Erin~C Mullany, Kalkidan~Hassen Abate, Cristiana
  Abbafati, Zegeye Abebe, Mohsen Afarideh, Anju Aggarwal, Sutapa Agrawal, Tomi
  Akinyemiju, Fares Alahdab, Umar Bacha, Victoria~F Bachman, Hamid Badali, Alaa
  Badawi, Isabela~M Bensenor, Eduardo Bernabe, Sibhatu Kassa~K Biadgilign,
  Stan~H Biryukov, Leah~E Cahill, Juan~J Carrero, Kelly~M. Cercy, Lalit
  Dandona, Rakhi Dandona, Anh~Kim Dang, Meaza~Girma Degefa, Maysaa {El Sayed
  Zaki}, Alireza Esteghamati, Sadaf Esteghamati, Jessica Fanzo, Carla~Sofia
  e~Sá~Farinha, Maryam~S Farvid, Farshad Farzadfar, Valery~L. Feigin, Joao~C
  Fernandes, Luisa~Sorio Flor, Nataliya~A. Foigt, Mohammad~H Forouzanfar,
  Morsaleh Ganji, Johanna~M. Geleijnse, Richard~F Gillum, Alessandra~C Goulart,
  Giuseppe Grosso, Idris Guessous, Samer Hamidi, Graeme~J. Hankey,
  Sivadasanpillai Harikrishnan, Hamid~Yimam Hassen, Simon~I. Hay, Chi~Linh
  Hoang, Masako Horino, Nayu Ikeda, Farhad Islami, Maria~D. Jackson, Spencer~L.
  James, Lars Johansson, Jost~B. Jonas, Amir Kasaeian, Yousef~Saleh Khader,
  Ibrahim~A. Khalil, Young-Ho Khang, Ruth~W Kimokoti, Yoshihiro Kokubo, G~Anil
  Kumar, Tea Lallukka, Alan~D Lopez, Stefan Lorkowski, Paulo~A. Lotufo, Rafael
  Lozano, Reza Malekzadeh, Winfried März, Toni Meier, Yohannes~A Melaku,
  Walter Mendoza, Gert~B.M. Mensink, Renata Micha, Ted~R Miller, Mojde
  Mirarefin, Viswanathan Mohan, Ali~H Mokdad, Dariush Mozaffarian, Gabriele
  Nagel, Mohsen Naghavi, Cuong~Tat Nguyen, Molly~R Nixon, Kanyin~L Ong,
  David~M. Pereira, Hossein Poustchi, Mostafa Qorbani, Rajesh~Kumar Rai,
  Christian Razo-García, Colin~D Rehm, Juan~A Rivera, Sonia
  Rodríguez-Ramírez, Gholamreza Roshandel, Gregory~A Roth, Juan Sanabria,
  Tania~G Sánchez-Pimienta, Benn Sartorius, Josef Schmidhuber,
  Aletta~Elisabeth Schutte, Sadaf~G. Sepanlou, Min-Jeong Shin, Reed~J.D.
  Sorensen, Marco Springmann, Lucjan Szponar, Andrew~L Thorne-Lyman, Amanda~G
  Thrift, Mathilde Touvier, Bach~Xuan Tran, Stefanos Tyrovolas, Kingsley~Nnanna
  Ukwaja, Irfan Ullah, Olalekan~A Uthman, Masoud Vaezghasemi, Tommi~Juhani
  Vasankari, Stein~Emil Vollset, Theo Vos, Giang~Thu Vu, Linh~Gia Vu, Elisabete
  Weiderpass, Andrea Werdecker, Tissa Wijeratne, Walter~C Willett, Jason~H Wu,
  Gelin Xu, Naohiro Yonemoto, Chuanhua Yu, and Christopher J~L Murray.
\newblock Health effects of dietary risks in 195 countries, 1990–2017: A
  systematic analysis for the {Global Burden of Disease Study 2017}.
\newblock {\em The Lancet}, 393(10184):1958--1972, 2019.

\bibitem{Zhao2020Dietary}
Zhiyun Zhao, Mian Li, Chao Li, Tiange Wang, Yu~Xu, Zhizheng Zhan, Weishan Dong,
  Zhiyong Shen, Min Xu, Jieli Lu, Yuhong Chen, Shenghan Lai, Wei Fan, Yufang
  Bi, Weiqing Wang, and Guang Ning.
\newblock Dietary preferences and diabetic risk in {China}: A large-scale
  nationwide {Internet} data-based study.
\newblock {\em Journal of Diabetes}, 12(4):270--278, 2020.

\bibitem{Joseph2009Nutrition}
James Joseph, Greg Cole, Elizabeth Head, and Donald Ingram.
\newblock Nutrition, brain aging, and neurodegeneration.
\newblock {\em The Journal of neuroscience : the official journal of the
  Society for Neuroscience}, 29(41):12795--12801, 2009.

\bibitem{Jensen2014NutriChem}
Kasper Jensen, Gianni Panagiotou, and Irene Kouskoumvekaki.
\newblock {{NutriChem}: A Systems Chemical Biology Resource to Explore the
  Medicinal Value of Plant-Based Foods}.
\newblock {\em Nucleic Acids Research}, 43(D1):D940--D945, 08 2014.

\bibitem{nian2021knowledge}
Yi~Nian, Jingcheng Du, Larry Bu, Fang Li, Xinyue Hu, Yuji Zhang, and Cui Tao.
\newblock Knowledge graph-based neurodegenerative diseases and diet
  relationship discovery.
\newblock {\em arXiv preprint arXiv:2109.06123}, 2021.

\bibitem{manica2019information}
Matteo Manica, Christoph Auer, Valery Weber, Federico Zipoli, Michele Dolfi,
  Peter Staar, Teodoro Laino, Costas Bekas, Akihiro Fujita, Hiroki Toda, et~al.
\newblock An information extraction and knowledge graph platform for
  accelerating biochemical discoveries.
\newblock {\em arXiv preprint arXiv:1907.08400}, 2019.

\bibitem{zhou2019application}
Lei Zhou, Chu Zhang, Fei Liu, Zhengjun Qiu, and Yong He.
\newblock Application of deep learning in food: {A} review.
\newblock {\em Comprehensive Reviews in Food Science and Food Safety},
  18(6):1793--1811, 2019.

\bibitem{Kiyoharu-FoodLog-CHSC2020}
Kiyoharu Aizawa.
\newblock Image recognition-based tool for food recording and analysis:
  Foodlog.
\newblock In {\em Connected Health in Smart Cities}, pages 1--9. 2020.

\bibitem{Chen-ZSIR-AAAI2020}
Jingjing Chen, Liangming Pan, Zhipeng Wei, Xiang Wang, ChongWah Ngo, and
  Tatseng Chua.
\newblock Zero-shot ingredient recognition by multi-relational graph
  convolutional network.
\newblock In {\em Proceedings of the {AAAI} Conference on Artificial
  Intelligence}, volume~34, pages 10542--10550, 2020.

\bibitem{Min-IGCMAN-ACMMM2019}
Weiqing Min, Linhu Liu, Zhengdong Luo, and Shuqiang Jiang.
\newblock {Ingredient-Guided Cascaded Multi-Attention Network} for food
  recognition.
\newblock In {\em Proceedings of {ACM} International Conference on Multimedia},
  page 1331–1339, 2019.

\bibitem{Min-MSMVFA-TIP2019}
Shuqiang Jiang, Weiqing Min, Linhu Liu, and Zhengdong Luo.
\newblock Multi-scale multi-view deep feature aggregation for food recognition.
\newblock {\em IEEE Transactions on Image Processing}, 29(1):265--276, 2019.

\bibitem{mezgec2017nutrinet}
Simon Mezgec and Barbara Korou{\v{s}}i{\'c}~Seljak.
\newblock {NutriNet}: A deep learning food and drink image recognition system
  for dietary assessment.
\newblock {\em Nutrients}, 9(7):657, 2017.

\bibitem{mezgec2019mixed}
Simon Mezgec, Tome Eftimov, Tamara Bucher, and Barbara~Korousic Seljak.
\newblock Mixed deep learning and natural language processing method for
  fake-food image recognition and standardization to help automated dietary
  assessment.
\newblock {\em Public Health Nutrition}, 22(7):1193--1202, 2019.

\bibitem{FangKLTC17}
Yuan Fang, Kingsley Kuan, Jie Lin, Cheston Tan, and Vijay Chandrasekhar.
\newblock Object detection meets knowledge graphs.
\newblock In {\em Proceedings of International Joint Conference on Artificial
  Intelligence}, pages 1661--1667, 2017.

\bibitem{DR2009Fuzzy}
D.R. Chittajallu, G.~Brunner, U.~Kurkure, R.P. Yalamanchili, and I.A.
  Kakadiaris.
\newblock {Fuzzy-Cuts}: A knowledge-driven graph-based method for medical image
  segmentation.
\newblock In {\em IEEE Conference on Computer Vision and Pattern Recognition},
  pages 715--722, 2009.

\bibitem{CHEN2012591}
Yufei Chen, Zhicheng Wang, Jinyong Hu, Weidong Zhao, and Qidi Wu.
\newblock The domain knowledge based graph-cut model for liver {CT}
  segmentation.
\newblock {\em Biomedical Signal Processing and Control}, 7(6):591--598, 2012.

\bibitem{s20154283}
Ya~Lu, Thomai Stathopoulou, Maria~F. Vasiloglou, Lillian~F. Pinault, Colleen
  Kiley, Elias~K. Spanakis, and Stavroula Mougiakakou.
\newblock {goFOODTM}: An artificial intelligence system for dietary assessment.
\newblock {\em Sensors}, 20(15), 2020.

\bibitem{9091215}
Ya~Lu, Thomai Stathopoulou, Maria~F. Vasiloglou, Stergios Christodoulidis, Zeno
  Stanga, and Stavroula Mougiakakou.
\newblock An artificial intelligence-based system to assess nutrient intake for
  hospitalised patients.
\newblock {\em {IEEE} Transactions on Multimedia}, 23:1136--1147, 2021.

\bibitem{ueland2020perspectives}
{\O}ydis Ueland, Themis Altintzoglou, Bente Kirkhus, Diana Lindberg,
  Guro~Helgesdotter Rogns{\aa}, Jan~Thomas Rosnes, Ida Rud, and Paula Varela.
\newblock Perspectives on personalised food.
\newblock {\em Trends in Food Science \& Technology}, 102:169--177, 2020.

\bibitem{kirk2021precision}
Daniel Kirk, Cagatay Catal, and Bedir Tekinerdogan.
\newblock Precision nutrition: A systematic literature review.
\newblock {\em Computers in Biology and Medicine}, 133:104365, 2021.

\bibitem{Chang-ontological-2008}
Chang-Shing Lee, Mei-Hui Wang, Huan-Chung Li, and Wen-Hui Chen.
\newblock Intelligent ontological agent for diabetic food recommendation.
\newblock In {\em IEEE International Conference on Fuzzy Systems}, pages
  1803--1810, June 2008.

\bibitem{Tumnark2013OntologyBasedPD}
Piyaporn Tumnark, Leandro Oliveira, and Nonchai Santibutr.
\newblock Ontology-based personalized dietary recommendation for weightlifting.
\newblock In {\em Proceedings of International Workshop on Computer Science in
  Sports}, 2013.

\bibitem{Vanesa-FQAS-SRND2013}
Vanesa Esp{\'i}n, Mar{\'i}a~V. Hurtado, Manuel Noguera, and Kawtar Benghazi.
\newblock Semantic-based recommendation of nutrition diets for the elderly from
  agroalimentary thesauri.
\newblock In {\em Flexible Query Answering Systems}, pages 471--482, 2013.

\bibitem{Chen-PFR-WSDM2021}
Yu~Chen, Ananya Subburathinam, Ching{-}Hua Chen, and Mohammed~J. Zaki.
\newblock Personalized food recommendation as constrained question answering
  over a large-scale food knowledge graph.
\newblock In {\em {Web Search and Data Mining}}, pages 544--552, 2021.

\bibitem{Azzi2020NutriSem}
Rabia Azzi, Sylvie Despres, and Gayo Diallo.
\newblock {NutriSem}: A semantics-driven approach to calculating nutritional
  value of recipes.
\newblock In {\em Trends and Innovations in Information Systems and
  Technologies}, pages 191--201, 2020.

\bibitem{shirai2021applying}
Sola Shirai, Oshani Seneviratne, and Deborah~L McGuinness.
\newblock Applying personal knowledge graphs to health.
\newblock {\em arXiv preprint arXiv:2104.07587}, 2021.

\bibitem{seneviratne2021personal}
Oshani Seneviratne, Jonathan Harris, Ching-Hua Chen, and Deborah~L McGuinness.
\newblock Personal health knowledge graph for clinically relevant diet
  recommendations.
\newblock {\em arXiv preprint arXiv:2110.10131}, 2021.

\bibitem{zhang2014guidance}
Jianrong Zhang and Tejas Bhatt.
\newblock A guidance document on the best practices in food traceability.
\newblock {\em Comprehensive Reviews in Food Science and Food Safety},
  13(5):1074--1103, 2014.

\bibitem{iqbal2017prospects}
Jamshed Iqbal, Zeashan~Hameed Khan, and Azfar Khalid.
\newblock Prospects of robotics in food industry.
\newblock {\em Food Science and Technology}, 37:159--165, 2017.

\bibitem{lee2006augmenting}
Chia-Hsun~Jackie Lee, Leonardo Bonanni, Jose~H Espinosa, Henry Lieberman, and
  Ted Selker.
\newblock Augmenting kitchen appliances with a shared context using knowledge
  about daily events.
\newblock In {\em Proceedings of the 11th international conference on
  Intelligent user interfaces}, pages 348--350, 2006.

\bibitem{Perona-VOV-IEEEPro2010}
P.~{Perona}.
\newblock Vision of a {Visipedia}.
\newblock {\em Proceedings of the {IEEE}}, 98(8):1526--1534, 2010.

\bibitem{Chen-NEIL-ICCV2013}
Xinlei Chen, Abhinav Shrivastava, and Abhinav Gupta.
\newblock {NEIL}: Extracting visual knowledge from web data.
\newblock In {\em IEEE International Conference on Computer Vision}, pages
  1409--1416, 2013.

\bibitem{Ferrari2007Learning}
Vittorio Ferrari and Andrew Zisserman.
\newblock Learning visual attributes.
\newblock In {\em Proceedings of International Conference on Neural Information
  Processing Systems}, pages 433--440, 2007.

\bibitem{Cewu-VRD-ECCV2016}
Cewu Lu, Ranjay Krishna, Michael~S. Bernstein, and Fei{-}Fei Li.
\newblock Visual relationship detection with language priors.
\newblock In {\em European Conference on Computer Vision}, volume 9905, pages
  852--869, 2016.

\bibitem{Xu-SGG-CVPR2017}
D.~{Xu}, Y.~{Zhu}, C.~B. {Choy}, and L.~{Fei-Fei}.
\newblock Scene graph generation by iterative message passing.
\newblock In {\em IEEE Conference on Computer Vision and Pattern Recognition},
  pages 3097--3106, 2017.

\bibitem{LiuMMKG-ESWC2019}
Ye~Liu, Hui Li, Alberto Garcia-Duran, Mathias Niepert, Daniel Onoro-Rubio, and
  David~S. Rosenblum.
\newblock {MMKG}: Multi-modal knowledge graphs.
\newblock In {\em The Semantic Web}, pages 459--474, 2019.

\bibitem{Papadopoulos-Pizza-CVPR2019}
Dim~P. Papadopoulos, Youssef Tamaazousti, Ferda Ofli, Ingmar Weber, and Antonio
  Torralba.
\newblock How to make a pizza: Learning a compositional layer-based {GAN}
  model.
\newblock In {\em IEEE/CVF Conference on Computer Vision and Pattern
  Recognition}, pages 7994--8003, 2019.

\bibitem{WangLHM20-SAGN-ECCV2020}
Hao Wang, Guosheng Lin, Steven C.~H. Hoi, and Chunyan Miao.
\newblock {Structure-aware Generation Network} for recipe generation from
  images.
\newblock In {\em European Conference on Computer Vision}, volume 12372, pages
  359--374, 2020.

\bibitem{Wu-CSGNN-TNNLS2020}
Zonghan Wu, Shirui Pan, Fengwen Chen, Guodong Long, Chengqi Zhang, and S~Yu
  Philip.
\newblock A comprehensive survey on graph neural networks.
\newblock {\em IEEE Transactions on Neural Networks and Learning Systems},
  32(1):4--24, 2021.

\bibitem{Park-FlavorGraph-SciRe2021}
Donghyeon Park, Keonwoo Kim, Seoyoon Kim, Michael Spranger, and Jaewoo Kang.
\newblock {FlavorGraph}: A large-scale food-chemical graph for generating food
  representations and recommending food pairings.
\newblock {\em Scientific Reports}, 11(1), 2021.

\bibitem{Tao2020Utilization}
Dandan Tao, Pengkun Yang, and Hao Feng.
\newblock Utilization of text mining as a big data analysis tool for food
  science and nutrition.
\newblock {\em Comprehensive Reviews in Food Science and Food Safety},
  19(2):875--894, 2020.

\bibitem{greenfield2003food}
Heather Greenfield and David~AT Southgate.
\newblock {\em Food Composition Data: Production, Management, and Use}.
\newblock Food \& Agriculture Org., 2003.

\bibitem{CHEN201654}
Shanquan Chen, Dandan Huang, Wenyan Nong, and Hoi~Shan Kwan.
\newblock Development of a food safety information database for {Greater
  China}.
\newblock {\em Food Control}, 65:54--62, 2016.

\bibitem{Hans2017Big}
Hans J.~P. Marvin, Esmée~M. Janssen, Yamine Bouzembrak, Peter J.~M.
  Hendriksen, and Martijn Staats.
\newblock Big data in food safety: An overview.
\newblock {\em Critical Reviews in Food Science and Nutrition},
  57(11):2286--2295, 2017.

\bibitem{BASHIARDES201857}
Stavros Bashiardes, Anastasia Godneva, Eran Elinav, and Eran Segal.
\newblock Towards utilization of the human genome and microbiome for
  personalized nutrition.
\newblock {\em Current Opinion in Biotechnology}, 51:57--63, 2018.

\bibitem{Nguyen-KGF-IF2020}
Hoang~Long Nguyen, Dang~Thinh Vu, and Jason~J. Jung.
\newblock Knowledge graph fusion for smart systems: A survey.
\newblock {\em Information Fusion}, 61:56--70, 2020.

\bibitem{Min2019A}
Weiqing Min, Shuqiang Jiang, Linhu Liu, Yong Rui, and Ramesh Jain.
\newblock A survey on food computing.
\newblock {\em {ACM} Computing Surveys}, 52(5):92:1--92:36, 2019.

\bibitem{Marino-UKGIC-CVPR2019}
K.~{Marino}, R.~{Salakhutdinov}, and A.~{Gupta}.
\newblock The more you know: Using knowledge graphs for image classification.
\newblock In {\em IEEE Conference on Computer Vision and Pattern Recognition},
  pages 20--28, 2017.

\bibitem{zhang2018variational}
Yuyu Zhang, Hanjun Dai, Zornitsa Kozareva, Alexander Smola, and Le~Song.
\newblock Variational reasoning for question answering with knowledge graph.
\newblock In {\em Proceedings of the {AAAI} Conference on Artificial
  Intelligence}, volume~32, pages 6069--6076, 2018.

\bibitem{Krieg-CookingAssistance2015}
Bernd Krieg-Br{\"u}ckner, Serge Autexier, Martin Rink, and Sidoine
  Ghomsi~Nokam.
\newblock {\em Formal Modelling for Cooking Assistance}, pages 355--376.
\newblock 2015.

\end{thebibliography}

\clearpage
\section*{Figure Captions}
\textbf{Figure 1 The evolution of the knowledge graph.}

This figure shows the development of main semantic data organizations above the arrow, from the semantic network to the knowledge graph. Below the arrow,  it displays key semantic web technologies. These web technologies are listed hierarchically, and each type of web technology relies on the capabilities of the layers below. With more technologies, more practical and powerful semantic data organization can be supported. The ultimate vision of semantic data organization is the Semantic Web, where all data are linked through relations.

\textbf{Figure 2 Pipeline of knowledge graph construction, representation, reasoning and applications.}

To construct a knowledge graph, a huge volume of data should be processed, including unstructured, semi-structured and structured data. Later, knowledge graphs can be constructed either manually or automatically, and the latter method mainly includes three components: knowledge extraction, knowledge fusion and knowledge refinement. Constructed knowledge graphs can be further used for representation learning and reasoning to support various tasks, such as search, recommendation and question answering.

\textbf{Figure 3 Flowchart of the study selection process.} 

We totally identified 167 studies from the databases. We removed 9 duplicate studies, excluded 83 irrelevant studies based on titles and abstracts. After manually adding several relevant studies, 83 studies are reviewed and evaluated in full for eligibility, and 58 studies meet all the criteria for this review and are thus included.

\textbf{Figure 4 A simplified structure of FoodOn~\cite{Dooley-FoodOn-npj2019}.}

In this example, the relations among food sources, products and related food processes of apples are described. Different entities are shown in different colors according to their classes. These entities are linked by different relations with different colors according to the type of relations.

\textbf{Figure 5 The structure of FoodKG~\cite{Steven-FoodKG-ISWC2019-1}.}

There are different instances of the FoodKG in the bottom left of the figure. FoodKG adopts the WhatToMake ontology as its ontology from several sources, such as FoodOn. Besides, instances in FoodKG are associated with nutrition data from the USDA ingredient nutrient database (the orange block at the top left) to support food recommendation with rich nutritional parameters.

\textbf{Figure 6 Applications of food knowledge graphs~(FKG).}

Representative applications of food knowledge graphs are shown: new recipe development, food question-answering systems, diet-diseases correlation discovery, visual food analysis, personalized dietary recommendation, food supply chain management, and food machinery intelligent manufacturing.

\end{document}